\documentclass[11pt]{article}
\usepackage[final]{acl}

\usepackage{times}
\usepackage{latexsym}
\usepackage{amsmath}
\usepackage{amssymb}
\usepackage{booktabs}
\usepackage{multirow}

\usepackage[T1]{fontenc}

\usepackage[utf8]{inputenc}

\usepackage{microtype}

\usepackage{inconsolata}

\usepackage{graphicx}

\usepackage{listings}
\usepackage{etoolbox}
\makeatletter
\@ifpackageloaded{lineno}{%
\AtBeginEnvironment{lstlisting}{\ifLineNumbers\nolinenumbers\let\acl@restorelno\linenumbers\else\let\acl@restorelno\relax\fi}
\AtEndEnvironment{lstlisting}{\acl@restorelno}
\AtBeginEnvironment{table}{\ifLineNumbers\nolinenumbers\let\acl@restorelno@tbl\linenumbers\else\let\acl@restorelno@tbl\relax\fi}
\AtEndEnvironment{table}{\acl@restorelno@tbl}
\AtBeginEnvironment{table*}{\ifLineNumbers\nolinenumbers\let\acl@restorelno@tbls\linenumbers\else\let\acl@restorelno@tbls\relax\fi}
\AtEndEnvironment{table*}{\acl@restorelno@tbls}
\AtBeginEnvironment{figure}{\ifLineNumbers\nolinenumbers\let\acl@restorelno@fig\linenumbers\else\let\acl@restorelno@fig\relax\fi}
\AtEndEnvironment{figure}{\acl@restorelno@fig}
\AtBeginEnvironment{figure*}{\ifLineNumbers\nolinenumbers\let\acl@restorelno@figs\linenumbers\else\let\acl@restorelno@figs\relax\fi}
\AtEndEnvironment{figure*}{\acl@restorelno@figs}
\AtBeginEnvironment{algorithm}{\ifLineNumbers\nolinenumbers\let\acl@restorelno@alg\linenumbers\else\let\acl@restorelno@alg\relax\fi}
\AtEndEnvironment{algorithm}{\acl@restorelno@alg}
}{\providecommand{\nolinenumbers}{}}
\makeatother
\lstdefinestyle{python}{
language=Python,
basicstyle=\ttfamily\scriptsize,
keywordstyle=\color{blue}\bfseries,
stringstyle=\color{red!70!black},
commentstyle=\color{green!50!black}\itshape,
showstringspaces=false,
numbers=none,
}
\usepackage[ruled,vlined]{algorithm2e}

\title{Revisiting Observation Reduction for Web Agents: \\Comprehensive Evaluation with a Lightweight Framework}

\author{Masafumi Enomoto, Ryoma Obara, Haochen Zhang\thanks{Work done while at NEC Corporation.}, Masafumi Oyamada \\
NEC Corporation \\
\texttt{\{masafumi-enomoto,ryoma-obara,haochen-zhang,oyamada\}@nec.com}}

\begin{document}
\maketitle
\begin{abstract}
HTML observations in LLM-based web agents are extremely long, and while many reduction methods have been proposed, it remains unclear which methods reduce overall agent latency while maintaining performance.
The main obstacle is the high cost of end-to-end evaluation: in our experiments, evaluating 11 methods across 32 configurations on 33 tasks of WorkArena L1 required 232.4 cumulative hours.
To address this, we propose a lightweight evaluation framework based on the Minimal Failure Set (MFS), the minimal set of HTML elements whose removal causes task failure.
We define coverage as the fraction of instances in which a reduction method fully retains the MFS, which serves as a proxy metric that requires neither web access nor LLM inference.
We validate that coverage strongly correlates with end-to-end success rate, with over 100$\times$ speedup in cumulative evaluation time on both benchmarks.
Using this framework, we find that extractive HTML reduction methods require either high computation cost or domain-specific optimization to reduce agent latency while maintaining performance.
Building on this, we optimize a pruning program on MFS training data, achieving 2.2$\times$ faster per-step latency on WorkArena L1 while retaining 84\% of the original success rate, and 3.1$\times$ faster on WebLinx while retaining 89\%.
\end{abstract}

\section{Introduction}
\label{sec:introduction}
LLM-based web agents have been actively studied in recent years, motivated by automating repetitive web tasks.
These agents sequentially observe web pages and issue actions to complete tasks, with observations typically represented as the DOM~\citep{related_work/mind2web, related_work/weblinx}, screenshots~\citep{related_work/autogui, related_work/seeclick}, or both~\citep{related_work/webvoyager, related_work/agent_s}.
While the DOM provides detailed information about web pages, raw HTML is extremely long; e.g., pages on the WorkArena benchmark~\citep{benchmark/workarena} range from 40K to 500K tokens, leading to high computation cost and latency of LLM inference.
To address this, various methods have been proposed to extract a relevant subset of HTML elements.
These methods vary in computation cost: program-based heuristics remove less relevant elements~\citep{browsergym}, retrieval-based approaches rank elements by relevance to the task~\citep{related_work/mind2web, related_work/html-t5, related_work/weblinx}, and LLM inference-based methods use an LLM to guide selection at higher cost~\citep{related_work/prune4web}.
Accessibility trees (a11y), which select interactive elements and canonicalize their representation, have also been adopted as an alternative heuristic~\citep{related_work/agent_occam, related_work/focusagent}.

Despite the variety of proposed methods, prior work lacks a comprehensive comparison across varying degrees of reduction, agent performance, and the computation cost and hence latency of the reduction step.
Critically, \textit{it remains unclear which methods reduce overall agent latency while maintaining performance, including the reduction cost itself.}
The main obstacle is the high cost of end-to-end evaluation, which requires feeding long observations to LLMs over multiple sequential steps.
Many benchmarks further require access to live web environments~\citep{benchmark/workarena, benchmark/webarena}, which introduces non-negligible latency.
In our experiments, evaluating 11 methods across 32 configurations on 33 tasks on WorkArena L1 with Qwen3.5-122B-A10B~\citep{models/qwen3} required 232.4 cumulative hours.
This cost hinders comprehensive comparison and development of observation reduction methods.

\begin{figure*}[t]
  \centering
  \includegraphics[width=1.0\linewidth]{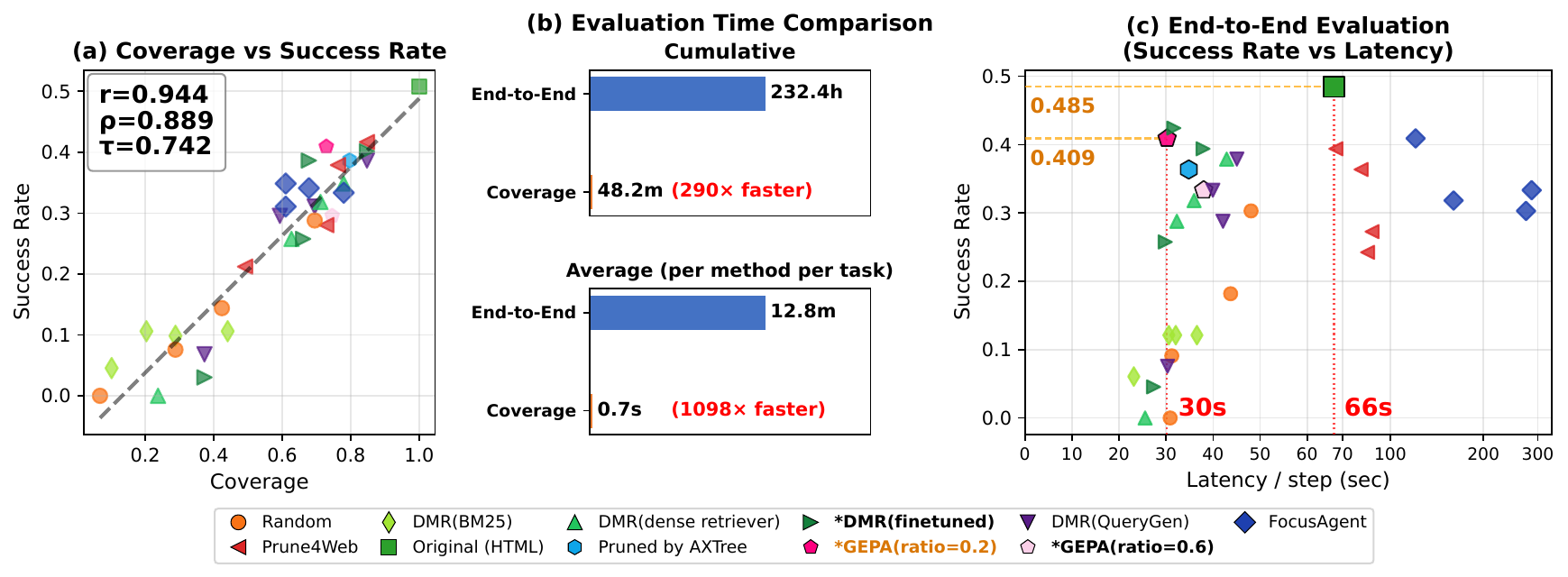}
  \caption{Key results of our framework on WorkArena L1 (33 tasks). (a) Coverage versus end-to-end success rate for each reduction method. (b) Evaluation time comparison between end-to-end evaluation with Qwen3.5-122B-A10B~\citep{models/qwen3} and coverage evaluation. (c) End-to-end success rate versus per-step latency when applying each reduction method, evaluated with Qwen3.5-122B-A10B. Methods marked with * are optimized on MFS training data.}
  \label{fig:killer}
\end{figure*}

To address this, we propose a lightweight evaluation framework for extractive HTML reduction methods.
The goal of any reduction method is to retain the elements necessary for task success while discarding redundant information.
This suggests a direct approach: for each HTML observation, pre-construct the minimal set of elements whose removal causes task failure, and use it as the evaluation criterion.
We call this set the \textbf{Minimal Failure Set (MFS)} and construct it via intervention experiments on successful trajectories on WorkArena L1~\citep{benchmark/workarena} and WebLinx~\citep{related_work/weblinx}.
We define \textbf{coverage} as the fraction of instances in which a reduction method fully retains the MFS, and propose it as a lightweight proxy for end-to-end task success rate.
Once MFSs are constructed, evaluating coverage requires neither policy model inference nor web access, and each instance can be evaluated independently in parallel.
We confirm that coverage strongly correlates with end-to-end success rate, with over 100$\times$ speedup in cumulative evaluation time on both benchmarks, as illustrated for WorkArena L1 in Figure~\ref{fig:killer}(a, b).

Using this framework, we comprehensively compare existing reduction methods across varying levels of HTML reduction and identify two key findings: (1) LLM inference-based methods achieve higher coverage than retrieval-based methods but at substantially higher latency, and this gap widens under aggressive reduction; (2) the HTML elements critical for task success differ across benchmarks: text content dominates on WebLinx, while HTML tags and attributes contribute more on WorkArena L1.
These findings suggest that extractive HTML reduction methods require either high computation cost or domain-specific optimization to reduce agent latency while maintaining performance.
Building on these, we optimize a pruning program for each benchmark on separate MFS training data using GEPA~\citep{gepa}, an LLM-based evolutionary optimization framework.
In end-to-end evaluation, this program achieves 2.2$\times$ faster per-step latency on WorkArena L1 while retaining 84\% of the original success rate (Figure~\ref{fig:killer}(c)), and 3.1$\times$ faster on WebLinx while retaining 89\%.
Domain-specific optimization of reduction programs thus emerges as the most effective approach on these benchmarks. Our framework supports this through coverage as both a fast evaluation metric and an optimization objective.

\section{A Lightweight Evaluation Framework for Observation Reduction Methods}
\label{sec:framework}
End-to-end evaluation of web agents is expensive, hindering both comprehensive comparison of reduction methods and automated search over configurations.
To address this, we construct the \textbf{Minimal Failure Set (MFS)}, the minimal set of HTML elements whose removal causes task failure, via intervention experiments on successful agent trajectories on each benchmark.
We define \textbf{coverage} as the fraction of instances in which a reduction method fully retains the MFS, and propose it as a lightweight proxy for end-to-end task success rate.
Once MFSs are constructed, evaluating coverage requires neither policy model inference nor web access, and each instance can be evaluated independently in parallel.

\subsection{Definitions}
\label{subsec:mfs_def}

\noindent\textbf{Minimal Failure Set.}
Let $H_s$ denote the HTML observation at step $s$ of an agent trajectory.
Each HTML element in $H_s$ is decomposed into a set of $(i, \mathit{attr})$ pairs, where $i$ is the element's unique identifier and $\mathit{attr}$ specifies the type of information.
These pairs serve as the unit of ablation: $\mathit{attr}$ can be a specific attribute name such as \texttt{value}, the special token \texttt{@tag} referring to the tag type itself, or \texttt{@text} referring to the element's direct text content.
For instance, given $\langle$\texttt{button value="OK"}$\rangle$\texttt{Submit}$\langle$\texttt{/button}$\rangle$ with identifier $42$, $(42, \texttt{value})$ refers to its \texttt{value} attribute, $(42, \texttt{@tag})$ refers to the tag \texttt{button}, and $(42, \texttt{@text})$ refers to the text \texttt{"Submit"}.
For a subset $X \subseteq H_s$ on a successful trajectory, we intervene by removing $X$ from $H_s$ at step $s$ and define $f(X) = 1$ if this causes task failure, and $f(X) = 0$ otherwise; any such $X$ with $f(X) = 1$ is called a \textbf{failure set}.
The smallest failure set is called the \textbf{Minimal Failure Set (MFS)}:
$X^* = \arg\min_{X \subseteq H_s,\, f(X)=1} |X|$.

\noindent\textbf{Coverage and Reduction Ratio.}
A reduction method $\mathcal{R}$ is a function that takes the full HTML $H_s$ as input
and returns a reduced HTML $\mathcal{R}(H_s) \subseteq H_s$ by selecting a subset of elements.
We evaluate $\mathcal{R}$ along two axes.
Let $\mathcal{D} = \{(H_s, \hat{X})\}$ be a set of instances derived from successful trajectories, each consisting of an HTML observation $H_s$ and an approximate MFS $\hat{X}$ constructed by the procedure described in \S\ref{subsec:mfs_construction}.
\textbf{Coverage} is the fraction of instances in $\mathcal{D}$ for which $\hat{X} \subseteq \mathcal{R}(H_s)$: $\mathrm{Coverage}(\mathcal{R}) = |\{d \in \mathcal{D} \mid \hat{X} \subseteq \mathcal{R}(H_s)\}| \,/\, |\mathcal{D}|$.
Note that a trivial method that retains the full HTML would achieve perfect coverage. We therefore also consider the \textbf{reduction ratio}: $\mathrm{RR}(\mathcal{R}) = \frac{1}{|\mathcal{D}|} \sum_{d \in \mathcal{D}} |\mathcal{R}(H_s)| \,/\, |H_s|$, where sizes are measured in characters; a smaller value indicates more aggressive reduction.
A method is more effective when it achieves high coverage at a low reduction ratio.

\subsection{Constructing Approximate MFSs}
\label{subsec:mfs_construction}

Finding the exact MFS $X^*$ would require up to $2^{|H_s|}$ interventions, each running the agent to task completion, which is infeasible.
We approximate $X^*$ in two phases: first, we use the agent's self-reported element references to prune the search space to a small candidate set; then, we apply iterative minimization to find a minimal subset.
As a prerequisite, we collect successful trajectories and record the agent's observations at each step.

\noindent\textbf{Trajectory Collection.}
We collect successful trajectories on the 33 tasks of WorkArena L1~\citep{benchmark/workarena} and 300 instances sampled from the test-iid split of WebLinx~\citep{related_work/weblinx}, using three LLMs:
Claude Sonnet 4.6~\citep{models/claude_sonnet_46},
Gemini 2.5 Flash~\citep{models/gemini_25}, and
GPT-5.1~\citep{models/gpt51},
to ensure diversity in the collected trajectories.
At each step $s$, we record: (1)~the HTML observation $H_s$, (2)~the set of $(i, \mathit{attr})$ pairs that the agent reports as relevant to its action, and (3)~the issued action.

\noindent\textbf{Phase 1: Intervention with Agent Self-Reports.}
For each sampled step $s$ from a successful trajectory, we test whether the self-reported elements are necessary for task success.
Let $E_s$ denote the set of $(i, \mathit{attr})$ pairs self-reported at step $s$.
For WorkArena, we reproduce the state at step $s$ by replaying the recorded actions on a fresh browser session.
Each recorded action refers to its target by an element identifier, but the DOM is not identical across sessions because of dynamic content such as timestamps and elements that depend on load timing.
We therefore normalize the DOM of both sessions, renumber the element identifiers sequentially, and match elements with the same identifier (Appendix~\ref{appendix:dom_normalization}).
Let $C_s \subseteq E_s$ denote the subset of self-reported pairs that actually exist in $H_s$.
We remove all pairs in $C_s$ from $H_s$ and run the agent from step $s$ to task completion.
If the task fails\footnote{For WebLinx, failure is defined as a step-wise reward score below $0.5$.} in both of two independent runs, $C_s$ serves as the candidate set for Phase~2.
We sample up to 10 steps per trajectory for WorkArena and all steps for WebLinx.

\noindent\textbf{Phase 2: Iterative Minimization.}
The candidate set $|C_s|$ has a mean size of 6.8 on WorkArena and 6.3 on WebLinx, so exhaustive search over all $2^{|C_s|}$ subsets remains costly.
To reduce the number of required tests, we apply the ddmin algorithm~\citep{ddmin}.
ddmin partitions $C_s$ into groups and tests whether each group can be removed while preserving failure. It repeats this at finer granularity and yields an approximate MFS $\hat{X} \subseteq C_s$ such that removing any single element from $\hat{X}$ no longer triggers failure.
For WorkArena, each ddmin test would require running the agent to task completion.
We instead use the erroneous action from Phase~1 as a proxy for failure judgment, reducing each test to a single inference step.
We also group elements by proximity in the DOM tree to further reduce the number of tests.
Details of the proxy oracle and the partitioning heuristic are described in Appendix~\ref{appendix:fps_ablation}.

\subsection{MFS Statistics and Construction Cost}
\label{subsec:mfs_dataset}

\noindent\textbf{MFS Statistics.}
On WorkArena, we generated 99 trajectories across three LLMs, of which 63 succeeded; 321 steps were sampled for Phase~1, 81 proceeded to Phase~2, and 59 valid MFS instances were obtained.
On WebLinx, we generated 900 trajectories, of which 329 succeeded; 129 steps were sampled, 44 proceeded to Phase~2, and 42 valid MFS instances were obtained.
Although the resulting MFS dataset is modest in size, its purpose is to rank reduction methods rather than to estimate task success directly.
We confirm in \S\ref{sec:exp_proxy/corr_w_end2end} that coverage strongly correlates with end-to-end success rate even with this modest number of MFS instances.
We also verify in Appendix~\ref{appendix:candidate_set} that restricting the candidate set to agent self-reported elements does not weaken this correlation.

The size of the approximate MFS, $|\hat{X}|$, averages 1.93 elements on WorkArena and 1.50 on WebLinx.
Most approximate MFSs are minimal: 80\% on WorkArena and 71\% on WebLinx consist of a single element.
For the remaining MFSs with multiple elements, the coverage metric requires all of them to be retained.

\noindent\textbf{Construction Cost.}
The pipeline required a total of 2{,}630 inferences on WorkArena and 1{,}296 on WebLinx, averaging 71K input and 379 output tokens per inference on WorkArena and 38K and 293 on WebLinx.
Coverage evaluation itself requires no model inference, so this one-time cost is amortized over subsequent evaluations of reduction method configurations.

\begin{figure*}[t]
  \centering
  \includegraphics[width=1.0\linewidth]{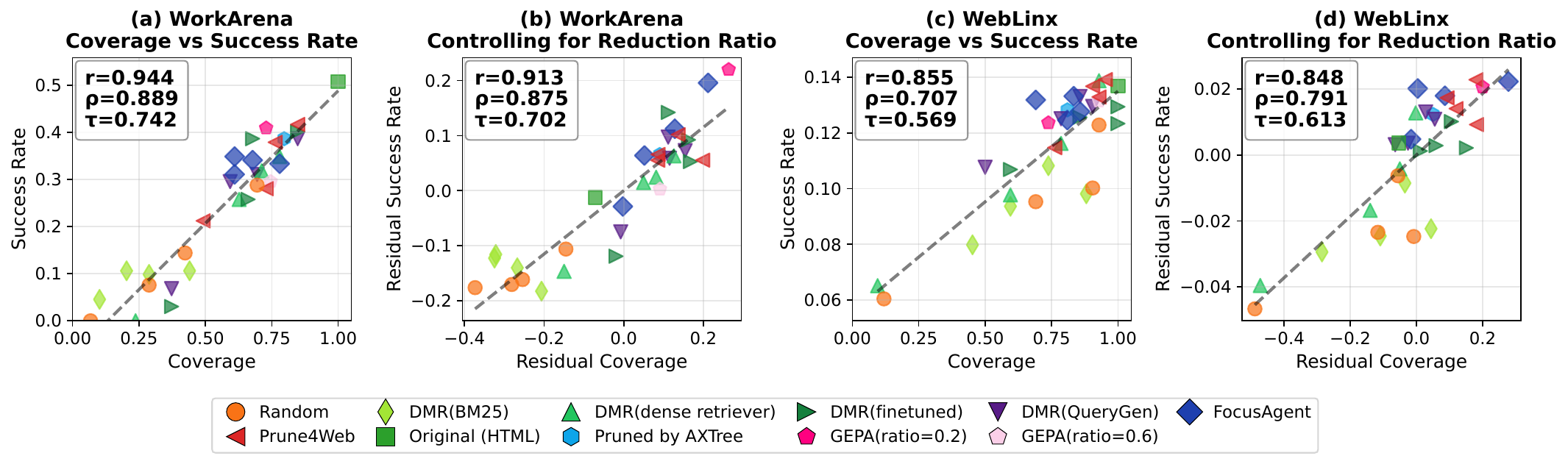}
  \caption{Correlation between coverage and end-to-end success rate across HTML reduction methods. (a, c) Scatter plots of mean coverage versus mean success rate for each (method, $k$) combination on WorkArena and WebLinx, respectively. (b, d) Partial correlation after regressing out reduction ratio from both coverage and success rate. Each panel reports Pearson's $r$, Spearman's $\rho$, and Kendall's $\tau$.}
  \label{fig:coverage_vs_success_rate}
\end{figure*}

\begin{table}[t]
\centering
\small
\caption{Evaluation time comparison between end-to-end and coverage evaluation. Cumulative runtime is the total wall-clock time to evaluate all reduction method configurations across all tasks. Average runtime is the mean wall-clock time per method per instance. WA and WL denote WorkArena L1 and WebLinx, respectively.}
\label{tab:timing_comparison}
\begin{tabular}{l|cc|cc}
\toprule
 & \multicolumn{2}{c|}{Cumulative} & \multicolumn{2}{c}{Average} \\
 & WA & WL & WA & WL \\
\midrule
Qwen3.5-122B & 232.4h & 117.0h & 12.8m & 42.6s \\
MiniMax-M2.5 & 272.2h & 136.2h & 15.1m & 49.5s \\
\midrule
Coverage & 48.2m & 28.5m & 0.7s & 0.6s \\
\bottomrule
\end{tabular}
\end{table}

\section{Coverage as a Fast Proxy}
\label{sec:exp_proxy}

\subsection{Experimental Setup}
\label{sec:exp_proxy/setup}

\noindent\textbf{Benchmarks.}
We evaluate on two web-agent benchmarks.
\textbf{WorkArena L1}~\citep{benchmark/workarena} consists of 33 task types on the ServiceNow platform,
requiring multi-step interaction with actual web services;
we use one seed per task type (33 tasks in total) to reduce evaluation cost.
\textbf{WebLinx}~\citep{related_work/weblinx} is a step-wise benchmark that evaluates
single-step action prediction; we use 300 instances sampled from the test-iid split.
All end-to-end success rates are averaged over two independent runs.
The MFS dataset for coverage evaluation is constructed from successful trajectories on these same tasks (\S\ref{subsec:mfs_construction}).

\noindent\textbf{Policy models.}
For end-to-end evaluation, we use two policy models:
Qwen3.5-122B-A10B~\citep{models/qwen3} and MiniMax-M2.5~\citep{models/minimax_m25}.

\noindent\textbf{Reduction methods.}
We evaluate the following observation reduction methods.
\textbf{Original (HTML)} passes the full HTML to the agent without any reduction, serving as an upper bound on information retention.
\textbf{Random} randomly selects $k$ elements from the HTML as a baseline.
\textbf{Pruned by AXTree} retains only HTML elements whose element identifiers appear in the accessibility tree (a11y).
\textbf{DMR (BM25)}~\citep{related_work/weblinx} treats each HTML element as a document by converting it to a structured text representation, then ranks elements by BM25 score against a query constructed from the task goal and action history, selecting the top $k$.
\textbf{DMR (Dense)}~\citep{related_work/weblinx} replaces BM25 with dense embedding retrieval (cosine similarity with Qwen3-Embedding-0.6B~\citep{qwen3emb}) using the same query.
\textbf{DMR (QueryGen)} extends DMR (Dense) by using an LLM to generate a search query before retrieval.
\textbf{FocusAgent}~\citep{related_work/focusagent} feeds the full HTML to an LLM and prompts it to select the $k$ most relevant elements by their HTML element identifiers.
\textbf{Prune4Web}~\citep{related_work/prune4web} uses a two-stage LLM pipeline: a Planner generates an action plan from a screenshot, and a Filter produces keyword weights. Elements are scored by a heuristic function and the top $k$ are selected.
For LLM inference-based methods, we use Qwen3.5-397B-A17B~\citep{models/qwen3}.
For methods with a selection parameter $k$, we evaluate $k \in \{10, 50, 100, 200\}$ for end-to-end evaluation and a wider range up to $k{=}500$ for coverage evaluation. Selected elements are expanded with ancestors, siblings, and descendants via a common tree-pruning post-process~\citep{related_work/mind2web}.
Implementation details including the tree-pruning procedure and prompts for each method are provided in Appendix~\ref{appendix:methods}.

To examine whether MFS data can serve as a training signal for reduction methods, we additionally evaluate methods optimized on separate benchmark-specific MFS training data (Appendix~\ref{appendix:training}).
\textbf{DMR (finetuned)} uses the same architecture as DMR (Dense) but fine-tunes the embedding model so that HTML elements containing MFS entries rank higher, using them as positive documents and sampling hard negatives from high-ranked non-MFS elements.
\textbf{GEPA (ratio=0.2)} and \textbf{GEPA (ratio=0.6)} are pruning programs optimized via GEPA~\citep{gepa}, an LLM-based evolutionary optimization framework. Each program is optimized to maximize coverage on the training data subject to the constraint that the reduction ratio stays below the target value (0.2 or 0.6). The seed and learned programs are presented in Appendix~\ref{appendix:gepa_programs}.

\subsection{Correlation with End-to-End Success Rate}
\label{sec:exp_proxy/corr_w_end2end}
Figure~\ref{fig:coverage_vs_success_rate} shows a strong correlation between coverage and end-to-end success rate on both WorkArena (panel~(a)) and WebLinx (panel~(c)), in terms of Pearson's $r$, Spearman's $\rho$, and Kendall's $\tau$.
Since retaining more HTML elements naturally increases both coverage and success rate, reduction ratio is a potential confound.
To control for this, we compute partial correlations by regressing out reduction ratio from both variables; the correlation remains strong on both benchmarks (panels~(b, d)).
This shows that methods with higher coverage achieve higher success rates even at the same reduction ratio.
The correlation is not driven by any specific reduction method: it remains strong after removing any single method, any method family, or the Original (HTML) configuration (Appendix~\ref{appendix:method_selection}).
Coverage therefore serves as a reliable metric for ranking the effectiveness of reduction methods, without requiring policy-model inference or web access.

\noindent\textbf{Sample Efficiency.}
The correlations reported above are computed from 59 MFS instances on WorkArena and 42 on WebLinx.
To verify that this modest number is sufficient, we vary the number of sampled instances and measure the rank correlation with end-to-end success rate.
The correlation rises steeply with the first few instances and plateaus early: on WorkArena, it exceeds $\rho = 0.7$ with only 4 instances; on WebLinx, $\rho = 0.7$ is reached with approximately 10 instances.
Since the correlation relies on a small number of MFS instances, we also check its reliability by bootstrap resampling of the instances: even at the lower end of the 95\% confidence interval, the correlation remains strong on WorkArena and moderate on WebLinx (Appendix~\ref{appendix:sample_eff}).
The full MFS dataset is thus sufficient for reliably evaluating reduction methods.

\noindent\textbf{Model Independence.}
Each MFS is derived from a trajectory of a specific LLM, raising the question of whether coverage is policy-dependent.
We compute coverage using MFS instances from each trajectory-collecting LLM (Gemini, Claude, GPT) separately and find that the correlations with end-to-end success rate are consistently high, ranging from 0.82 to 0.88 on WorkArena and 0.66 to 0.71 on WebLinx.
Since the end-to-end success rate is measured with policy models distinct from the trajectory-collecting LLMs, this suggests that MFS captures information necessary for solving the task across models, rather than information specific to any single LLM's trajectory.
Details are provided in Appendix~\ref{appendix:model_ind}.

\noindent\textbf{Generalization to Unseen Tasks.}
Coverage also predicts the success rate on tasks that are not used to construct the MFS instances.
On WorkArena, we split the 33 task types into two disjoint groups and compute the correlation between coverage on one group and the end-to-end success rate on the other.
Over 50 random splits, Spearman's $\rho$ is 0.838 on average.
On WebLinx, coverage computed on MFS instances from tasks that are not used in the end-to-end evaluation also correlates with the success rate.
Details are provided in Appendix~\ref{appendix:unseen_tasks}.

\subsection{Evaluation Time Comparison}
\label{sec:exp_proxy/eval_time_comparison}
Table~\ref{tab:timing_comparison} compares evaluation time between end-to-end and coverage evaluation in terms of cumulative runtime and average runtime per instance, where the latter serves as a lower bound under maximum parallelization.
On WorkArena, coverage evaluation achieves a 290$\times$ speedup in cumulative runtime and a 1098$\times$ speedup in average runtime compared to end-to-end evaluation with Qwen3.5-122B.
On WebLinx, the speedup is 246$\times$ and 71$\times$, respectively; the smaller per-instance speedup reflects the fact that WebLinx evaluates single-step action prediction without live web access.
Together, these results confirm that coverage is a fast and reliable proxy for evaluating observation reduction methods.

\begin{table}[t]
\centering
\small
\caption{Average wall-clock time (seconds) to reduce a single HTML observation, measured during the coverage experiments.}
\label{tab:reduction_latency}
\begin{tabular}{lrr}
\toprule
Method & WorkArena & WebLinx \\
\midrule
\textit{Baseline} \\
\quad Original (HTML) & 0.01 & 0.00 \\
\quad Random & 0.10 & 0.03 \\
\midrule
\textit{Program-based} \\
\quad Pruned by AXTree & 0.27 & 0.04 \\
\quad GEPA (ratio=0.6) & 0.10 & 0.03 \\
\quad GEPA (ratio=0.2) & 0.10 & 0.11 \\
\midrule
\textit{Retrieval-based} \\
\quad DMR (BM25) & 0.12 & 0.04 \\
\quad DMR (Dense) & 2.95 & 1.13 \\
\quad DMR (finetuned) & 2.55 & 1.06 \\
\midrule
\textit{LLM inference} \\
\quad DMR (QueryGen) & 13.79 & 8.79 \\
\quad Prune4Web & 28.32 & 20.14 \\
\quad FocusAgent & 105.69 & 78.71 \\
\bottomrule
\end{tabular}
\end{table}

\begin{figure*}[t]
  \centering
  \includegraphics[width=1.0\linewidth]{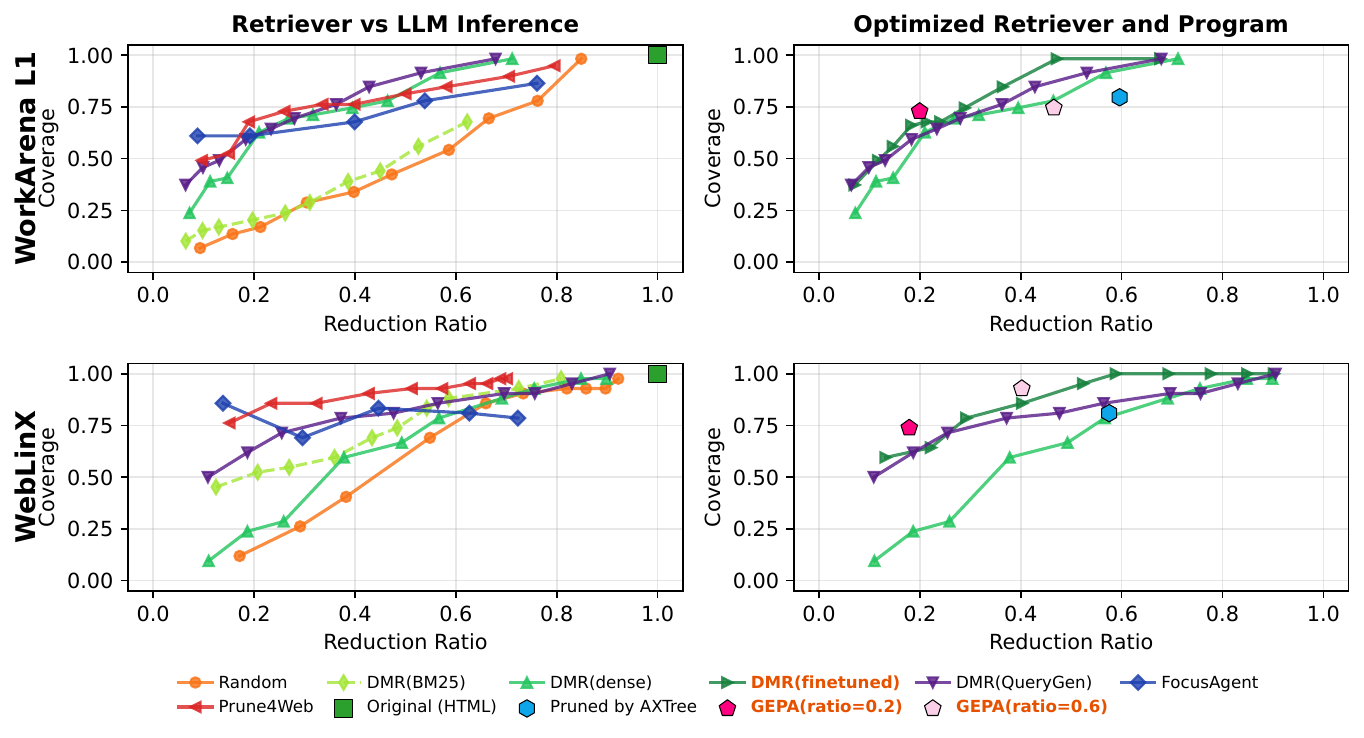}
  \caption{Coverage versus reduction ratio on WorkArena L1 (top) and WebLinx (bottom). Left column compares retrieval-based and LLM inference-based methods alongside baselines. Right column shows the effect of MFS-based optimization: DMR (finetuned) is compared against DMR (Dense) and DMR (QueryGen) as a reference from the LLM inference group, while GEPA is compared against Pruned by AXTree as an unoptimized program-based method.}
  \label{fig:coverage_and_latency}
\end{figure*}

\section{Comprehensive Comparison of Reduction Methods}
\label{sec:comp_comparison}
We utilize coverage to comprehensively compare existing reduction methods across a wide range of configurations. While higher computation cost in the reduction step may improve coverage, it incurs additional latency that may offset the faster policy model inference from reduced input. We thus evaluate each method along both coverage and latency.

\subsection{Comparison using Coverage}
\label{sec:comp_comparison/with_coverage}

\noindent\textbf{Trade-off between latency and performance.}
Table~\ref{tab:reduction_latency} reports the average wall-clock time for each method to reduce a single HTML observation.
Latency increases with the computational approach: program-based methods complete in under 1 second, retrieval-based methods require a few seconds, and LLM inference-based methods take over 10 seconds.
Figure~\ref{fig:coverage_and_latency} (left column) compares the coverage of retrieval-based and LLM inference-based methods alongside baselines.
Among methods not optimized on MFS data, LLM inference-based methods achieve higher coverage than retrieval-based methods, but at much higher latency: for example, FocusAgent requires over 100 seconds per observation on WorkArena, while retrieval-based methods complete in a few seconds.
At low reduction ratios, this performance gap is large and LLM inference-based methods are clearly superior.
Conversely, as the reduction ratio increases, the gap narrows and even the Random baseline approaches the coverage of more expensive methods.
The trade-off between latency and performance is thus most relevant when aggressive reduction is required.

\noindent\textbf{Need for domain-specific optimization.}
The relative ranking of methods varies across benchmarks: for instance, BM25 performs comparably to the dense retriever on WebLinx but underperforms on WorkArena.
To investigate this, we ablate individual HTML element types and measure the resulting coverage drop (Table~\ref{tab:html_elem_ablation}).
The results reveal that the elements critical for task success differ between the two benchmarks.
On WebLinx, removing text content causes a large coverage drop (59.5\%), while individual tags and attributes have comparatively small effects.
In contrast, WorkArena relies more heavily on CSS-related attributes such as \texttt{id} and \texttt{class}, which rank among the top contributors to coverage drop.
BM25 is favorable on WebLinx because text content dominates relevance, but fails on WorkArena where attributes matter more.
This suggests that methods without training are not guaranteed to perform well across benchmarks, highlighting the need for domain-specific optimization.

\noindent\textbf{Training on MFS data improves low-latency methods.}
Figure~\ref{fig:coverage_and_latency} (right column) compares methods optimized on MFS data with their unoptimized counterparts.
GEPA (ratio=0.2, 0.6), a program-based method optimized on MFS data, achieves lower reduction ratios than Pruned by AXTree without sacrificing coverage.
DMR (finetuned), which fine-tunes the dense retriever on MFS data, improves over DMR (Dense) and approaches the Pareto frontier of LLM inference-based methods such as DMR (QueryGen).
Since both methods only optimize a retriever or a pruning program, their latency remains low, demonstrating that MFS data can be used to improve reduction performance of low-latency methods.

\begin{figure*}[t]
  \centering
  \includegraphics[width=\linewidth]{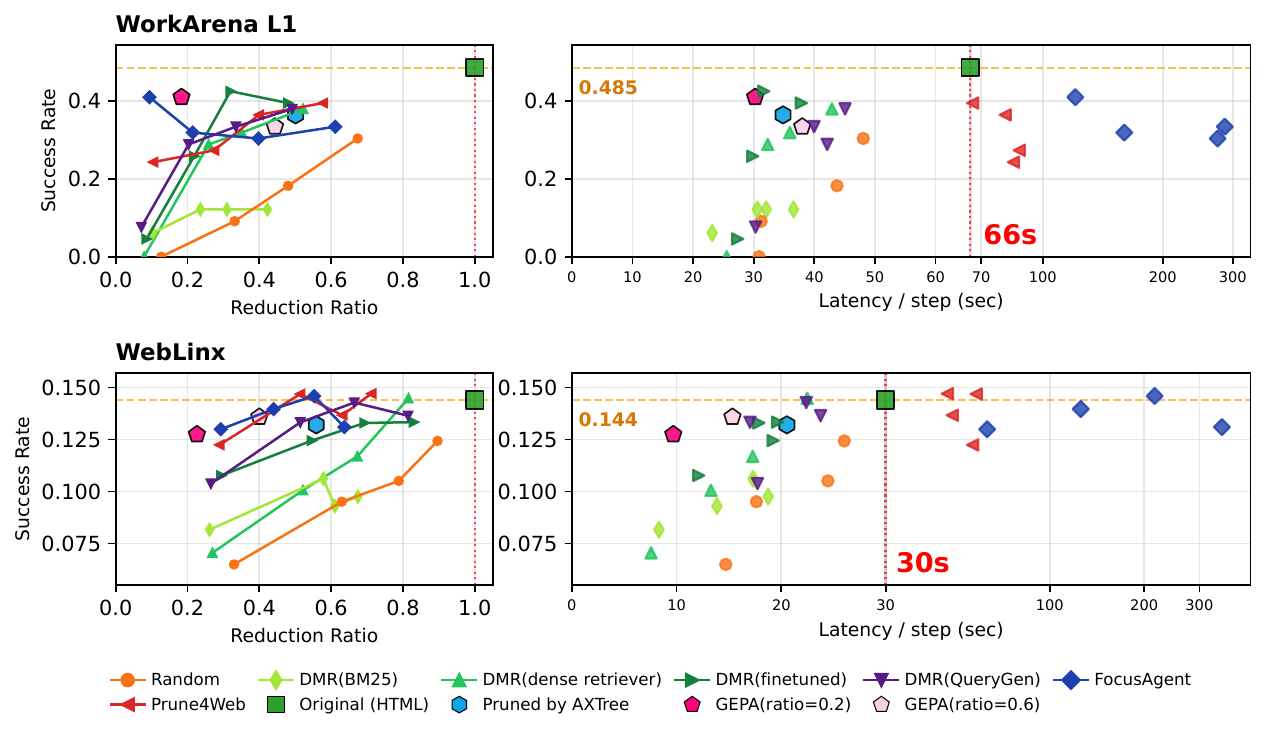}
  \caption{End-to-end evaluation with Qwen3.5-122B-A10B as the policy model on WorkArena L1 (top) and WebLinx (bottom). Left column: success rate versus reduction ratio. Right column: success rate versus average wall-clock latency per step, which includes reduction and policy model inference; WorkArena additionally includes web access latency. The red dashed line and orange dashed line indicate the latency and success rate of passing the full HTML without reduction, respectively.}
  \label{fig:end2end}
\end{figure*}

\begin{table}[t]
\centering
\small
\caption{Coverage drop (\%) when ablating a single HTML element type. Top-5 tags and attributes ranked by coverage drop are shown for each benchmark.}
\label{tab:html_elem_ablation}
\begin{tabular}{l|lr|lr}
\toprule
 & \multicolumn{2}{c|}{WorkArena L1} & \multicolumn{2}{c}{WebLinx} \\
\midrule
Text & & 30.5 & & 59.5 \\
\midrule
\multirow{5}{*}{Tag} & select & 13.6 & a & 7.1 \\
 & span & 8.5 & button & 4.8 \\
 & a & 6.8 & div & 4.8 \\
 & td & 5.1 & input & 2.4 \\
 & input & 5.1 & h2 & 2.4 \\
\midrule
\multirow{5}{*}{Attr} & value & 22.0 & href & 16.7 \\
 & id & 16.9 & value & 7.1 \\
 & class & 10.2 & class & 7.1 \\
 & selected & 5.1 & aria-label & 7.1 \\
 & onclick & 3.4 & data-testid & 4.8 \\
\bottomrule
\end{tabular}
\end{table}

\subsection{End-to-End Evaluation}
\label{sec:comp_comparison/end2end}
In this section, we conduct end-to-end evaluation combining each reduction method with a policy model. We examine whether the ranking of methods is consistent with the coverage evaluation, and measure the total latency per step, which includes reduction and policy model inference; WorkArena additionally includes web access latency.

\noindent\textbf{Trend consistency with coverage evaluation.}
Figure~\ref{fig:end2end} (left column) shows the success rate versus reduction ratio. Among methods not optimized on MFS data, LLM inference-based methods achieve relatively higher success rates than retrieval-based methods, and the performance gap narrows as the reduction ratio increases. These trends are consistent with the coverage evaluation (\S\ref{sec:comp_comparison/with_coverage}).

\noindent\textbf{MFS-optimized reduction programs retain performance.}
GEPA (ratio=0.2) achieves a reduction ratio of approximately 0.2 while limiting the success rate drop from the original HTML.
At this reduction ratio, it matches or exceeds the performance of all other methods: in paired tests, it is significantly better in 5 of the 12 comparisons and never significantly worse (Appendix~\ref{appendix:paired}).
DMR (finetuned) improves over DMR (Dense) overall.
However, like other retrieval-based methods, its performance drops sharply at low reduction ratios on WorkArena.
A case study in Appendix~\ref{appendix:case_studies} shows one cause: retrieval-based methods miss an interactive element whose text shares no word with the task goal, whereas GEPA retains it by a structural rule.
Training on MFS data does not remove this drop, likely because of a difference in optimization objectives: GEPA directly maximizes coverage under a target reduction ratio constraint, whereas DMR (finetuned) only trains the retriever to rank HTML elements containing MFS entries higher.

\noindent\textbf{MFS-optimized reduction programs reduce latency.}
Figure~\ref{fig:end2end} (right column) shows the success rate versus total latency per step. Since GEPA (ratio=0.2) is program-based, its reduction latency is negligible, and the total per-step latency is substantially reduced. On WorkArena, it reduces latency from 65.7s to 30.2s (2.2$\times$ faster) while retaining 84\% of the original success rate. On WebLinx, it achieves a 3.1$\times$ latency reduction while retaining 89\% of the success rate.
In contrast, LLM inference-based methods achieve comparable success rates but offer limited latency reduction because the reduction step itself requires substantial inference time; FocusAgent and Prune4Web exceed the latency of the no-reduction baseline in some configurations.
A breakdown of the per-step latency shows that the speedup of GEPA (ratio=0.2) comes from faster policy model inference on the shorter input, whereas LLM inference-based methods spend most of the time in the reduction step (Appendix~\ref{appendix:latency_decomp}).
Overall, MFS-optimized program-based methods offer the best balance of performance and latency under aggressive reduction.
Results with MiniMax-M2.5 as the policy model show consistent trends (Appendix~\ref{appendix:minimax}).

\section{Related Work}
\label{sec:related_work}
\noindent\textbf{Observation Reduction in Web Agents.}
Observations in LLM-based web agents are typically represented as the DOM~\citep{related_work/mind2web, related_work/html-t5}, screenshots~\citep{related_work/autogui, related_work/seeclick, related_work/clickagent}, or both~\citep{related_work/webvoyager, related_work/agent_s, related_work/omniparser}.
While the DOM provides detailed information about web pages, raw HTML is extremely long, which leads to high computation cost and latency.
To address this, various extractive reduction methods have been proposed.
Retrieval-based approaches score HTML elements against a query derived from the task instruction and action history, and select the most relevant ones~\citep{related_work/mind2web, related_work/html-t5, related_work/weblinx}.
Program-based heuristics remove less relevant elements such as \texttt{script} and \texttt{meta} tags~\citep{browsergym}.
LLM inference-based methods use an LLM to guide selection; \citet{related_work/prune4web} generate keyword weights via an LLM and score elements with a heuristic function.
Accessibility trees (a11y) provide a more compact representation of the DOM by extracting interactive elements and canonicalizing their representation.
\citet{related_work/agent_occam} use program-based heuristics to simplify the a11y by merging redundant elements and compacting structured content, and \citet{related_work/focusagent} prompt an LLM to directly output the line numbers of task-relevant elements.
\citet{related_work/lcow} go beyond selection by adding natural-language descriptions to extracted elements to enhance decision-making.

\section{Conclusion}
\label{sec:conclusion}
We proposed a lightweight evaluation framework for observation reduction methods based on the Minimal Failure Set (MFS).
We defined coverage based on MFS and validated it as a reliable proxy for end-to-end success rate, achieving over 100$\times$ speedup in cumulative evaluation time.
Our comprehensive comparison revealed that extractive HTML reduction methods require either high computation cost or domain-specific optimization to reduce agent latency while maintaining performance.
Building on this finding, we optimized a pruning program on MFS training data, achieving 2.2$\times$ faster per-step latency with Qwen3.5-122B-A10B on WorkArena L1 while retaining 84\% of the original success rate, and 3.1$\times$ faster on WebLinx while retaining 89\%.

\section*{Limitations}
\noindent\textbf{Scope of Coverage.}
Coverage is designed for extractive methods that select a subset of HTML elements, and cannot evaluate representation-transforming methods such as summarization or semantic compression. Two methods that retain the same elements but differ in output representation would achieve identical coverage.

\noindent\textbf{Trajectory Dependence.}
We construct MFS instances from successful trajectories of three diverse LLMs (Gemini 2.5 Flash, Claude Sonnet 4.6, and GPT-5.1), and verify that coverage computed from any single LLM's trajectories correlates similarly with end-to-end success rate (Appendix~\ref{appendix:model_ind}).
Nevertheless, MFS can only reflect trajectories that were actually observed, and elements required by unobserved trajectories are not captured.

\noindent\textbf{Data Availability.}
We do not redistribute the MFS instances because of restrictions on the underlying HTML observations.
For WorkArena, the HTML pages are from ServiceNow non-production instances whose terms of use prohibit redistribution.\footnote{\url{https://www.servicenow.com/terms-of-use.html}}
For WebLinx, the original dataset is publicly available under CC BY-NC-SA 4.0,\footnote{\url{https://huggingface.co/datasets/McGill-NLP/WebLINX}} but since the HTML pages originate from third-party websites, we refrain from redistributing derivatives containing the raw HTML.
We provide the MFS construction procedure, reduction method implementations, and evaluation settings in the paper and appendix to enable reconstruction.

\bibliography{main}
\appendix
\nolinenumbers
\raggedbottom
\setcounter{dbltopnumber}{3}
\renewcommand{\dbltopfraction}{0.85}
\renewcommand{\dblfloatpagefraction}{0.75}
\setcounter{topnumber}{3}
\renewcommand{\topfraction}{0.85}
\renewcommand{\textfraction}{0.1}
\renewcommand{\floatpagefraction}{0.75}

\section{Implementation Details of Reduction Methods}
\label{appendix:methods}

\subsection{Query Construction for DMR}
\label{appendix:dmr_query}

\noindent\textbf{Query string.}
For DMR (BM25), DMR (Dense), and DMR (QueryGen), the retrieval query is constructed
by concatenating the task goal and the action history in the following format:

\begin{lstlisting}
Goal: Create a new change request with short description "Network issue"

Previous Actions:
- Step 0: fill('a196', 'CHG0000013')
- Step 1: fill('a671', 'Ernest piquance...')
- Step 2: click('a340')
\end{lstlisting}

\noindent\textbf{Element text representation.}
Each HTML element is converted to a structured text representation in WebLinx format~\citep{related_work/weblinx}:

\begin{lstlisting}
[[tag]] button
[[xpath]] /html/body/div/form/button
[[bid]] a585
[[text]] Submit Form
[[attributes]] class='btn-primary' id='submit-btn' role='button'
[[children]] span
\end{lstlisting}

The \texttt{[[text]]} field is truncated to 200 characters; each attribute value in \texttt{[[attributes]]} is truncated to 100 characters; \texttt{[[children]]} lists at most 5 child tag names.
The following attributes are included: \texttt{class}, \texttt{id}, \texttt{name}, \texttt{role}, \texttt{aria-label}, \texttt{placeholder}, \texttt{value}, \texttt{href}, \texttt{title}, \texttt{type}, \texttt{for}, \texttt{src}, \texttt{alt}, \texttt{data-testid}.

\noindent\textbf{BM25 retriever.}
Standard BM25 with parameters $k_1 = 1.5$, $b = 0.75$.
Elements are ranked by BM25 score against the query; the top $k$ are selected.

\noindent\textbf{Dense retriever.}
Query and element representations are embedded using an embedding model and ranked by cosine similarity.
We use Qwen3-Embedding-0.6B~\citep{qwen3emb} as the embedding model, served via an OpenAI-compatible local API.

\subsection{DMR (QueryGen) Prompt}
\label{appendix:dmr_querygenp}

DMR (QueryGen) uses an LLM to generate a higher-quality search query before dense retrieval.
The system prompt is as follows:

\begin{lstlisting}
You are a query generator for a dense retrieval system that finds relevant DOM elements.

## Task
Generate a search query to retrieve DOM elements needed for the next action in a web agent task.

## Retrieval System

Each DOM element is converted to text in the following format and embedded:

- `[[tag]]`: HTML tag name
- `[[xpath]]`: XPath to the element
- `[[bid]]`: Unique element identifier
- `[[text]]`: Direct text content of the element (max 200 chars)
- `[[attributes]]`: Important attributes (class, id, name, role, aria-label, placeholder, value, href, title, type, for, src, alt, data-testid)
- `[[children]]`: Child element tag names

Your query is embedded and compared against all element embeddings using cosine similarity. Consider this representation format when generating your query.

## Output Format
<think>
[Analyze the task goal and action history to determine what DOM element the agent needs to interact with next]
</think>
<query>
[Your search query]
</query>
\end{lstlisting}

The user message is constructed as follows:

\begin{lstlisting}
# Task
{goal}

# Action History
{action_history}

Generate a search query to find the DOM element needed for the next action.
\end{lstlisting}

The generated query is extracted from \texttt{<query>...</query>} tags.

\subsection{FocusAgent Prompt}
\label{appendix:focusagent_prompt}

FocusAgent~\citep{related_work/focusagent} was originally proposed for accessibility tree observations.
In this work, we adapt it to HTML by prompting the LLM to select $k$ bids from the full HTML directly.
The system prompt is as follows:

\begin{lstlisting}
You are part of a web agent whose job is to solve a task. You
are currently at a step of the whole episode, and your job is
to extract the relevant information for solving the task. An
agent will execute the task after you on the subset that you
extracted. Make sure to extract sufficient information to be
able to solve the task, but also remove information that is
irrelevant to reduce the size of the observation and all the
distractions.
\end{lstlisting}

The user prompt instructs the model to select exactly $k$ elements:

\begin{lstlisting}
# Instructions
Select the TOP {k} most relevant elements for the task at this step of completion.
A final HTML will be built from these elements. It should contain
enough information to understand the state of the page, the
current step and to perform the right next action, including
buttons, links and any element to interact with.

You MUST select exactly {k} elements, ranked by relevance.
If there are fewer than {k} relevant elements, include less relevant
ones to reach {k}. If there are more than {k} relevant elements,
select only the {k} most important ones.

Expected answer format:
<think>
Reason about which {k} elements of the HTML are most important to achieve
the goal specified in # Goal. Rank them by relevance.
</think>
<answer>
A list of exactly {k} bid values of the most relevant elements, ranked by importance.
For example: [1, 24, 35, 47, 123]
</answer>

# Goal:
{goal}

# History of interaction with the task:
{history}

# Observation:
{html_txt}
\end{lstlisting}

The response is parsed from \texttt{<answer>...</answer>} tags.
Selected bids are then passed to the tree-pruning post-process described in Appendix~\ref{appendix:tree_pruning}.

\subsection{Prune4Web Pipeline}
\label{appendix:prune4web}

We follow the two-stage pipeline of \citet{related_work/prune4web} with minor adaptations.

\noindent\textbf{Stage 1: Planner.}
The Planner receives the task goal, action history, available action space, and a screenshot image.
Its system prompt is as follows:

\begin{lstlisting}
You are a planning agent that breaks down tasks into executable UI steps with strict safety protocols. Follow ABSOLUTELY:

Core Rules:
1. POP-UP HANDLING: Only close non-normal pop-ups that block or interfere with the task (e.g., ads, mandatory login walls, cookie consent dialogs). Do not close or dismiss any normal UI pop-ups that do not affect task execution (e.g., search suggestion dropdowns, informational tooltips).
2. UI-ACTION FORMATTING: Phrase steps as EXACT interface commands using ONLY the actions defined in the Action space below.
3. LOGIN RESTRICTIONS: NEVER trigger login UNLESS task explicitly mentions credentials or an undismissable login wall appears.
4. TERMINATION CRITERIA: TERMINATE SOLELY when: login wall appears WITH NO close option, paywall or other physical UI blockage occurs, or system security could be compromised.

{action_space}

Output Format:
{
  "state_analysis": "Brief context analysis",
  "progress_evaluation": "X% - Description",
  "challenges": ["list"],
  "next_steps": ["Only output ONE action using the Action space format above"],
  "action_type": "<action name from Action space above>",
  "target": {
    "text": "ELEMENT TEXT to interact with (e.g., 'Search box', 'Login button')"
  },
  "reasoning": "Security/UI rationale"
}

Critical Instructions:
- "action_type" must be one of the actions defined in the Action space (e.g., "click", "fill", "select_option")
- "target.text" must ONLY contain the TEXT OF THE ELEMENT to interact with
- For "fill" actions: "target.text" is the element to fill (e.g., "Search box"), NOT the value to fill
- The value to fill should ONLY appear in "next_steps", NOT in "target.text"
\end{lstlisting}

The user message contains a screenshot image, the task goal, and the action history.
The Planner outputs a JSON action plan including \texttt{action\_type} and \texttt{target.text}.

\noindent\textbf{Stage 2: Filter.}
The Filter receives the Planner's JSON output. Its system prompt is as follows:

\begin{lstlisting}
You are a professional filter keyword generator. Your task is to generate keywords with their corresponding weights for filtering and scoring interactive elements based on the complete plan output from the Planner. Output the thinking process in <think> ... </think> tags, and the final answer in <answer> ... </answer> tags.

Task Description:
Analyze the Planner's output and generate relevant keywords with appropriate weights that can be used to score and filter webpage elements for the given task.

Keyword Weighting Strategy:
- Assign higher weights (e.g., 30-50) to critical, task-specific keywords.
- Assign medium weights (e.g., 10-25) for supporting or contextual terms.
- Assign lower weights (e.g., 1-10) for general relevance terms.

Output Format:
<think>
[Your keyword analysis and weight assignment thinking process]
</think>
<answer>
{
  "keyword_weights": {
    "keyword1": weight1,
    "keyword2": weight2
  }
}
</answer>
\end{lstlisting}

The user message is the Planner's JSON output passed verbatim.
The Filter outputs a keyword weight dictionary such as
\texttt{\{"Search": 40, "input": 30, \ldots\}},
with higher weights assigned to task-critical terms.

\noindent\textbf{Scoring.}
The scoring function computes a relevance score for each HTML element by checking whether the Filter's keywords appear in the element's attributes.
For each element, display text ($\beta = 1.0$), semantic attributes such as \texttt{aria-label}, \texttt{placeholder}, \texttt{name}, and \texttt{role} ($\beta = 0.8$), and structural attributes such as \texttt{class} and \texttt{id} ($\beta = 0.5$) are extracted, where $\beta$ is the attribute weight.
For each keyword-attribute pair, the function tries four match types in priority order and assigns a match quality $\alpha$: exact string match ($\alpha = 1.0$), phrase containment ($\alpha = 0.8$), stemmed word match ($\alpha = 0.6$), and fuzzy string similarity ($\alpha = 0.4 \times \text{similarity}$).
Let $K$ denote the set of keywords from the Filter, with $w_k$ the weight of keyword $k$, and $A(e)$ the attributes of element $e$.
The element score is:
\begin{align}
  \text{Score}(e) = \sum_{a \in A(e)} \sum_{k \in K} w_k \cdot \alpha(k, a) \cdot \beta_a
\end{align}
Listing~\ref{lst:prune4web_scoring} shows the implementation.
Elements are ranked by score and the top $k$ are selected.

\begin{lstlisting}[caption={Prune4Web scoring function.},label={lst:prune4web_scoring},style=python]
import re
from rapidfuzz import fuzz
from nltk.stem import PorterStemmer

stemmer = PorterStemmer()

def normalize(text):
    return re.sub(r'\s+', ' ', text.lower()).strip()

def fuzzy_score(keyword, text, tokens):
    return max(
        fuzz.partial_ratio(keyword, text) / 100,
        max((fuzz.ratio(keyword, t) / 100
             for t in tokens), default=0.0))

def score_element(element, keyword_weights):
    tiers = [
        (element.text,        1.0),  # Tier 1
        (element.aria_label,  0.8),  # Tier 2
        (element.placeholder, 0.8),
        (element.name,        0.8),
        (element.role,        0.8),
        (element.class_attr,  0.5),  # Tier 3
        (element.id_attr,     0.5),
    ]
    score = 0.0
    for attr_text, beta in tiers:
        if not attr_text:
            continue
        t = normalize(attr_text)
        tokens = t.split()
        stemmed = [stemmer.stem(w) for w in tokens]
        for kw, w in keyword_weights.items():
            k = normalize(kw)
            if t == k:
                alpha = 1.0           # exact
            elif ' ' in k and k in t:
                alpha = 0.8           # phrase
            elif stemmer.stem(kw) in stemmed:
                alpha = 0.6           # word
            else:
                fs = fuzzy_score(k, t, tokens)
                if fs >= 0.75:
                    alpha = 0.4 * fs  # fuzzy
                else:
                    continue
            score += w * alpha * beta
    return score
\end{lstlisting}

\noindent\textbf{Adaptations from the original paper.}
We add \texttt{name} and \texttt{role} to Tier 2 attributes, which are not explicitly specified in the original paper.

\subsection{Training Details for MFS-Optimized Methods}
\label{appendix:training}

DMR (finetuned) and GEPA are trained on MFS data constructed from agent trajectories that are disjoint from the evaluation data used in \S\ref{sec:framework}.
For WorkArena L1, we use the same 33 task types with different random seeds, yielding 81 MFS instances.
For WebLinx, we use instances from the test-iid split excluding the 300 instances used for evaluation, yielding 249 MFS instances.
The MFS construction procedure follows the same protocol described in \S\ref{subsec:mfs_construction}.

\paragraph{DMR (finetuned).}
DMR (finetuned) fine-tunes the Qwen3-Embedding-0.6B model~\citep{qwen3emb} used by DMR (Dense) so that MFS elements rank higher given the same query.
Each MFS element is treated as a positive document and the remaining HTML elements as negatives.
Hard negatives are sampled from the top-5 to top-20 ranked elements (by cosine similarity with the base model) that are not positives, up to 7 per positive.
We train with MultipleNegativesRankingLoss, a learning rate of 1e-5, warmup ratio 0.1, fp16 precision, and early stopping with patience 3.
The batch size is 32 for WorkArena (105 training tuples) and 16 for WebLinx (244 training tuples).
Each fine-tuned model is served on a single NVIDIA RTX PRO 6000 GPU via TEI (Text Embeddings Inference).

\paragraph{GEPA.}
We use GEPA~\citep{gepa}, an LLM-based evolutionary optimization framework, to optimize pruning programs.
We use Gemini 2.5 Flash~\citep{models/gemini_25} as the mutation LLM, with a mini-batch size of 5, 8 parallel workers, and a 50\% train/validation split (41/40 instances for WorkArena, 125/124 for WebLinx).
A separate program is optimized for each target reduction ratio $r_{\text{target}} \in \{0.2, 0.6\}$.
Let $H'_i$ denote the reduced HTML for instance $i$, $H_i$ the original HTML, and $M_i$ the set of MFS elements.
The per-instance score is defined as:
\begin{equation}
  s_i = \mathbf{1}\left[\frac{|H'_i|}{|H_i|} \leq r_{\text{target}}\right] \cdot \mathbf{1}\left[M_i \subseteq H'_i\right],
\end{equation}
where $\mathbf{1}[\cdot]$ is the indicator function and $|{\cdot}|$ denotes the character length. The score is 1 only when the reduced output satisfies both the size constraint and full MFS retention; otherwise it is 0.
The overall objective is $\frac{1}{N}\sum_{i=1}^{N} s_i$.
The full optimization takes 70 calls to the mutation LLM and 1.2 hours of wall-clock time on WorkArena, and 114 calls and 2.7 hours on WebLinx.
The seed program and learned programs are presented in Appendix~\ref{appendix:gepa_programs}.

\subsection{Tree Pruning Post-Process}
\label{appendix:tree_pruning}

All element-selection methods (DMR variants, FocusAgent, Prune4Web)
apply a common tree-pruning post-process after element selection,
based on the implementation of \citet{related_work/mind2web}.
Given the set of selected bids, the pruner retains:

\begin{itemize}
  \item the selected elements themselves,
  \item all ancestor elements up to the root,
  \item descendants up to depth 5 (at most 50 children per node),
  \item up to 3 preceding and following siblings.
\end{itemize}

Non-retained elements are unwrapped (their children are moved to their parent before deletion),
so that the structural context of selected elements is preserved without introducing disconnected subtrees.

\textbf{Pruned by AXTree} uses a stricter configuration: descendants are retained only up to depth 1,
and no siblings are retained (\texttt{max\_sibling = 0}).
This reflects the fact that the accessibility tree itself provides a compact, structurally pruned view,
so additional context expansion is unnecessary.

\section{Robustness Analysis of Coverage}
\label{appendix:mfs_sampling}

This appendix addresses four questions about coverage as a proxy for end-to-end success rate.
Appendix~\ref{appendix:method_selection} tests whether the correlation depends on a particular reduction method.
Appendix~\ref{appendix:sample_model} tests whether the number of MFS instances is sufficient and whether the result depends on the LLM that collected the trajectories.
Appendix~\ref{appendix:unseen_tasks} tests whether coverage predicts the success rate on tasks that are not used to construct the MFS instances.
Appendix~\ref{appendix:candidate_set} tests whether limiting the MFS candidates to agent self-reports weakens the correlation.

\subsection{Robustness to the Choice of Reduction Methods}
\label{appendix:method_selection}

\begin{table*}[t]
\centering
\small
\caption{Spearman's $\rho$ between coverage and end-to-end success rate when methods or configurations are removed. Leave-one-method-out and leave-one-family-out report the minimum and maximum over the removed choices. Partial denotes the partial correlation after regressing out reduction ratio.}
\label{tab:leave_out}
\begin{tabular}{lcccc}
\toprule
 & \multicolumn{2}{c}{WorkArena L1} & \multicolumn{2}{c}{WebLinx} \\
 & $\rho$ & Partial $\rho$ & $\rho$ & Partial $\rho$ \\
\midrule
All 32 configurations & 0.889 & 0.875 & 0.707 & 0.791 \\
Leave-one-method-out & [0.852, 0.923] & [0.831, 0.909] & [0.621, 0.784] & [0.738, 0.871] \\
Leave-one-family-out & [0.841, 0.921] & [0.839, 0.883] & [0.632, 0.819] & [0.711, 0.841] \\
Excluding the full HTML point & 0.878 & 0.850 & 0.686 & 0.813 \\
\bottomrule
\end{tabular}

\end{table*}

The correlations in \S\ref{sec:exp_proxy/corr_w_end2end} are computed over 32 configurations of 11 reduction methods.
We group these methods into four families as in Table~\ref{tab:reduction_latency}: baseline, program-based, retrieval-based, and LLM inference-based methods.
To test whether the correlation depends on the choice of reduction methods, we remove one method with all its $k$ settings, or one family, and recompute the correlation.
We also remove the Original (HTML) configuration, which has a coverage of 1 and the highest success rate.
Table~\ref{tab:leave_out} shows that Spearman's $\rho$ stays between 0.841 and 0.923 on WorkArena and between 0.621 and 0.819 on WebLinx.
The correlation therefore does not depend on any particular reduction method or family.

\subsection{Sample Efficiency and Model Independence}
\label{appendix:sample_model}

We investigate (1)~how many MFS instances are needed for coverage to reliably rank reduction methods, and (2)~whether the choice of trajectory-collecting LLM affects the ranking.

\begin{figure*}[t]
  \centering
  \includegraphics[width=\linewidth]{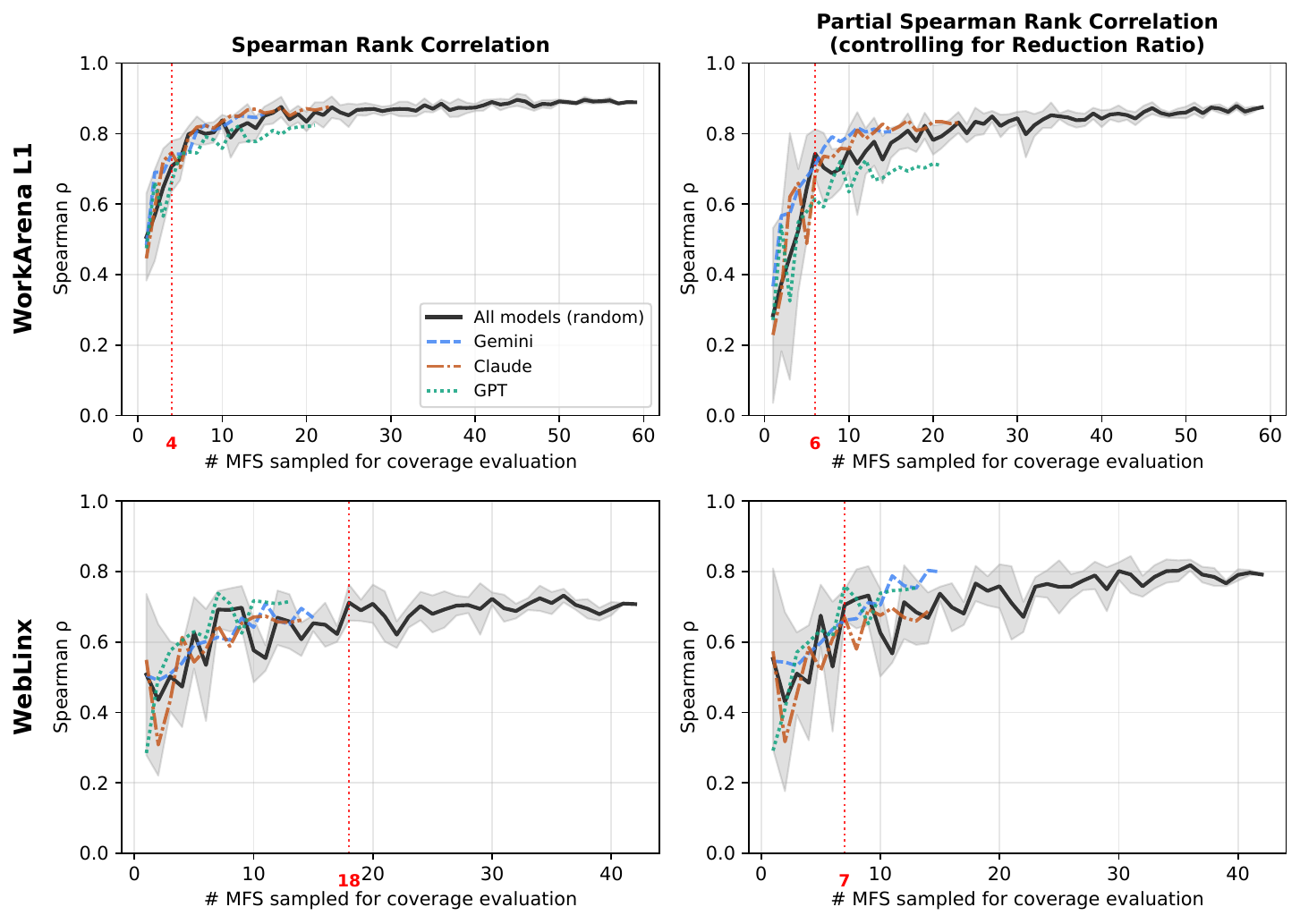}
  \caption{Spearman rank correlation between coverage and end-to-end success rate as a function of the number of MFS instances sampled. Left column: raw correlation; right column: partial correlation controlling for reduction ratio. Top row: WorkArena L1; bottom row: WebLinx. Solid black lines show sampling from all trajectory-collecting LLMs; gray bands indicate $\pm$1 standard deviation across 5 random sampling trials. Dashed colored lines show LLM-specific subsets. Red vertical lines mark the smallest $N$ where the mean correlation first exceeds 0.7.}
  \label{fig:mfs_sampling}
\end{figure*}

\noindent\textbf{Setup.}
Each MFS instance consists of an HTML observation $H_s$ and its approximate MFS $\hat{X}$, derived from a successful trajectory collected by one of three LLMs: Gemini 2.5 Flash, Claude Sonnet 4.6, and GPT-5.1.
For WorkArena L1, 59 instances are available (15 from Gemini, 23 from Claude, 21 from GPT); for WebLinx, 42 instances (15/14/13).
We vary the number of sampled MFS instances $N$ from 1 to the maximum.
For each $N$, we compute the mean coverage for each of the 32 reduction method configurations and measure the Spearman rank correlation with the end-to-end success rate.
The end-to-end success rate for each configuration is averaged over two policy models (Qwen3.5-122B-A10B and MiniMax-M2.5), yielding 32 data points in total.
We also compute partial Spearman rank correlation after regressing out the reduction ratio.
The sampling is repeated 5 times per $N$ and the correlations are averaged.

\subsubsection{Sample Efficiency}
\label{appendix:sample_eff}
Figure~\ref{fig:mfs_sampling} shows that coverage evaluation is remarkably sample-efficient.
On WorkArena, the Spearman rank correlation exceeds $\rho = 0.7$ with only 4 instances; on WebLinx, $\rho = 0.7$ is reached with approximately 10 instances.
The full MFS datasets (59 and 42 instances) are thus sufficient for reliably evaluating reduction methods.

\begin{table*}[t]
\centering
\small
\caption{Bootstrap 95\% confidence intervals of the correlation between coverage and end-to-end success rate. Partial denotes the partial correlation after regressing out reduction ratio from both variables.}
\label{tab:corr_ci}
\begin{tabular}{lccc}
\toprule
 & Pearson $r$ & Spearman $\rho$ & Kendall $\tau$ \\
\midrule
WorkArena L1 & 0.944 [0.895, 0.951] & 0.889 [0.827, 0.913] & 0.742 [0.661, 0.774] \\
\quad Partial & 0.913 [0.837, 0.925] & 0.875 [0.769, 0.886] & 0.702 [0.569, 0.718] \\
WebLinx & 0.855 [0.761, 0.888] & 0.707 [0.557, 0.786] & 0.569 [0.428, 0.628] \\
\quad Partial & 0.848 [0.720, 0.883] & 0.791 [0.585, 0.847] & 0.613 [0.427, 0.677] \\
\bottomrule
\end{tabular}

\end{table*}

Since this correlation relies on a small number of MFS instances, we quantify its uncertainty by bootstrap resampling: we resample the 59 instances on WorkArena and the 42 instances on WebLinx with replacement 10,000 times.
For each resample, we recompute the coverage of every configuration and compute its correlation with the end-to-end success rate.
Table~\ref{tab:corr_ci} shows the resulting 95\% confidence intervals.
Even at the lower ends of these intervals, Spearman's $\rho$ is 0.827 on WorkArena and 0.557 on WebLinx, so the correlation remains strong on WorkArena and moderate on WebLinx.

\subsubsection{Model Independence}
\label{appendix:model_ind}
When MFS data is restricted to trajectories collected by a single LLM (Gemini-only, Claude-only, or GPT-only), the resulting correlations converge to similar values, as shown by the dashed lines in Figure~\ref{fig:mfs_sampling}.
On WorkArena, the correlation using all available instances from each LLM ranges from 0.82 to 0.88; on WebLinx, from 0.66 to 0.71.
This indicates that the correlation between coverage and end-to-end success rate is not driven by the behavioral or self-reporting patterns of any particular trajectory-collecting LLM.

\subsection{Generalization to Unseen Tasks}
\label{appendix:unseen_tasks}

\begin{figure*}[t]
  \centering
  \includegraphics[width=\linewidth]{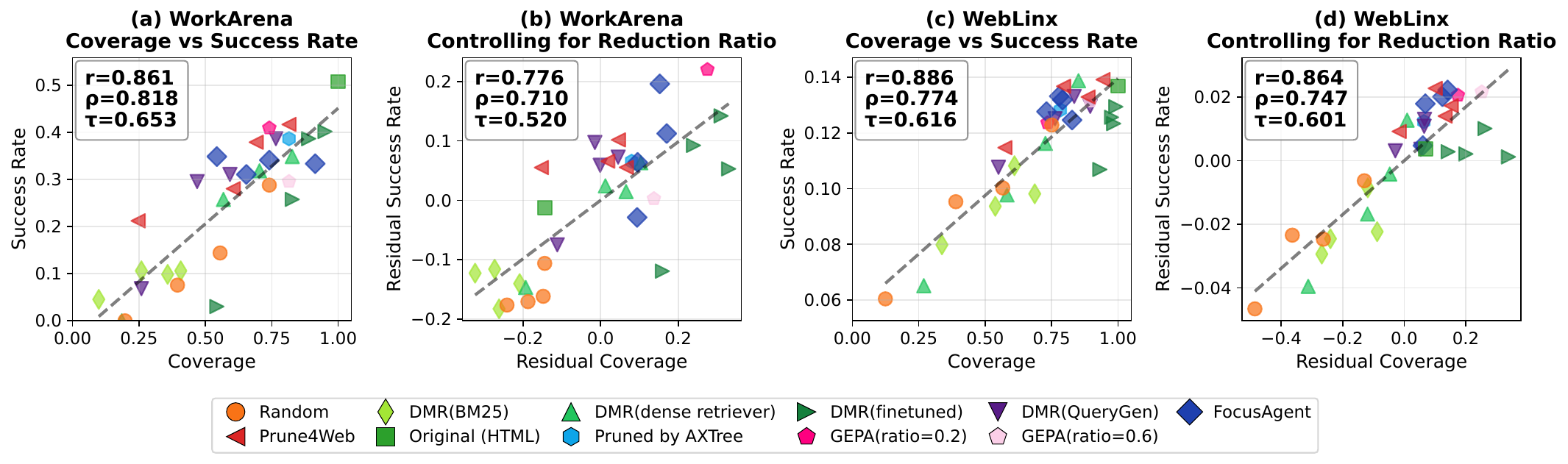}
  \caption{Correlation between coverage and end-to-end success rate, with coverage computed on separately constructed MFS instances. (a, c) Scatter plots of mean coverage versus mean success rate for each (method, $k$) combination on WorkArena and WebLinx, respectively. (b, d) Partial correlation after regressing out reduction ratio. Each panel reports Pearson's $r$, Spearman's $\rho$, and Kendall's $\tau$.}
  \label{fig:coverage_correlation_train}
\end{figure*}

\noindent\textbf{Separately Constructed MFS.}
The coverage results in \S\ref{sec:exp_proxy/corr_w_end2end} use MFS instances derived from the same tasks as the end-to-end evaluation.
To test whether MFS constructed from other tasks also serves as a valid proxy, we compute coverage on a separate set of MFS instances, constructed in the same way as in \S\ref{subsec:mfs_construction}: 81 instances on WorkArena\footnote{The WorkArena MFS training data uses the same 33 task types with different random seeds, so this tests robustness to seed variation rather than to unseen task types.} and 249 on WebLinx (Appendix~\ref{appendix:training}).
Figure~\ref{fig:coverage_correlation_train} shows that the correlation remains strong: Spearman's $\rho = 0.818$ on WorkArena and $\rho = 0.774$ on WebLinx, with partial correlations of $\rho = 0.710$ and $\rho = 0.747$.

\begin{table}[t]
\centering
\small
\caption{Correlation between coverage and end-to-end success rate on WorkArena L1 when coverage and success rate are computed on disjoint groups of task types. For each of 50 random splits, coverage is computed on one group and the success rate on the other, and then the two groups are swapped. The statistics are over the resulting 100 correlations.}
\label{tab:unseen_tasks}
\setlength{\tabcolsep}{2.8pt}
\begin{tabular}{lccc}
\toprule
 & All task types & \multicolumn{2}{c}{Disjoint split} \\
\cmidrule(lr){3-4}
 & & Mean $\pm$ s.d. & [Min, max] \\
\midrule
Pearson $r$ & 0.944 & 0.886 $\pm$ 0.028 & [0.796, 0.936] \\
Spearman $\rho$ & 0.889 & 0.838 $\pm$ 0.041 & [0.748, 0.920] \\
Kendall $\tau$ & 0.742 & 0.688 $\pm$ 0.051 & [0.597, 0.795] \\
\bottomrule
\end{tabular}

\end{table}

\noindent\textbf{Disjoint Task Types on WorkArena.}
The analysis above uses the same task types on WorkArena.
To test whether coverage predicts the success rate on unseen task types, we randomly split the 33 task types of WorkArena L1 into two disjoint groups $A$ and $B$.
For each of the 32 (method, $k$) configurations, we compute coverage using only the MFS instances from $A$ and the end-to-end success rate using only the tasks in $B$.
We then compute the correlation over these 32 pairs.
We repeat this for 50 random splits in both directions, which yields 100 correlations.
Table~\ref{tab:unseen_tasks} shows that the correlation remains strong: Spearman's $\rho$ is 0.838 on average and 0.748 even for the worst split.
Together, these results suggest that MFS captures which HTML elements are critical within a web domain, rather than task-specific information, allowing coverage to function as a proxy across different sets of tasks.

\subsection{Robustness to Candidate Set Restriction}
\label{appendix:candidate_set}

Our MFS construction restricts the candidate set to agent self-reported elements (\S\ref{subsec:mfs_construction}), excluding elements that the agent did not report as relevant.
To test whether this restriction biases coverage as a proxy, we reconstruct MFS from an expanded candidate set on WorkArena L1.
The expanded set augments the self-reported elements with three additional sources: BM25 retrieval, dense retrieval, and DOM-adjacent elements of the action target.
BM25 top-5 and dense retrieval top-5 are obtained using the same retriever configurations as DMR (BM25) and DMR (Dense) in \S\ref{sec:exp_proxy/setup}.
DOM-adjacent elements include up to 5 structural neighbors of the HTML element specified in the agent's action at step $s$: parent, children, siblings, and ancestors up to 2 levels.
Phase~1 and Phase~2 are re-run from scratch using this expanded set on the trajectories of the MFS training data (Appendix~\ref{appendix:training}), yielding 83 valid MFS instances.

\begin{figure*}[t]
  \centering
  \includegraphics[width=\linewidth]{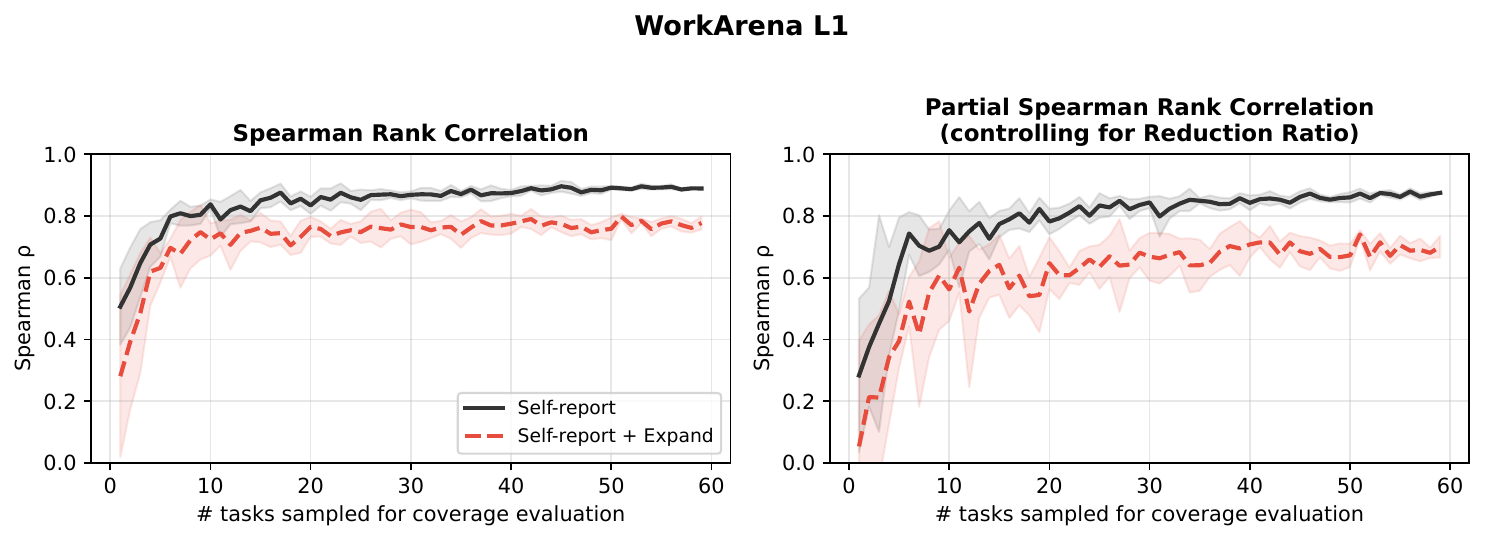}
  \caption{Spearman rank correlation between coverage and end-to-end success rate on WorkArena L1, comparing self-report MFS and expanded MFS. Left: raw correlation. Right: partial correlation controlling for reduction ratio. Gray and red bands indicate $\pm$1 standard deviation across 5 random sampling trials.}
  \label{fig:mfs_selfreport_robustness}
\end{figure*}

Figure~\ref{fig:mfs_selfreport_robustness} compares the subsampling curves of self-report MFS and expanded MFS.
Both achieve strong Spearman rank correlation with end-to-end success rate, but expanded MFS does not improve over self-report MFS: at $N{=}59$, self-report MFS reaches $\rho = 0.89$ while expanded MFS reaches $\rho = 0.78$.
The same trend holds for partial correlation controlling for reduction ratio ($\rho = 0.87$ vs.\ $\rho = 0.70$).
Since ddmin returns one 1-minimal set among potentially many, a larger candidate set may cause ddmin to converge on elements outside the agent's self-report.
The lower correlation suggests that such elements are less informative for ranking reduction methods.
These results confirm that the self-report restriction does not undermine coverage as a proxy, and that agent self-reports provide a sufficient basis for constructing MFS that reliably ranks reduction methods.

\section{Policy Model Serving}
\label{appendix:serving}

For end-to-end evaluation, policy models (Qwen3.5-122B-A10B, Qwen3.5-397B-A17B-FP8, and MiniMax-M2.5) are served via vLLM~(v0.17.0) on a single node with 8$\times$ NVIDIA A100 GPUs using 8-way tensor parallelism.
The maximum context length is 131{,}072 tokens.

\section{Details of ddmin in MFS Construction}
\label{appendix:fps_ablation}

\subsection{Background: The ddmin Algorithm}

The ddmin algorithm~\citep{ddmin} is a general-purpose test input minimization algorithm.
Given an oracle $\text{test}(S)$ that returns FAIL or PASS for a subset $S \subseteq C$, ddmin finds a 1-minimal subset $M \subseteq C$ such that $\text{test}(M) = \text{FAIL}$ and $\text{test}(M \setminus \{c\}) = \text{PASS}$ for every $c \in M$.

As shown in Algorithm~\ref{alg:ddmin}, the algorithm iteratively partitions $C$ into $n$ chunks and tests whether each chunk can be removed (i.e., whether $\text{test}(C \setminus \Delta_i) = \text{FAIL}$).
If removing a chunk preserves failure, $C$ is reduced to $C \setminus \Delta_i$ and $n$ is decreased by one.
If no chunk can be removed, $n$ is doubled.
The algorithm terminates when $n \geq |C|$, at which point every individual element has been tested and the result is 1-minimal.
The number of oracle calls is $O(\log |C|)$ in the best case and $O(|C|^2)$ in the worst case.

\begin{algorithm}[t]
\caption{ddmin algorithm~\citep{ddmin}}\label{alg:ddmin}
\KwIn{Failure-inducing set $C$, oracle $\text{test}$}
\KwOut{1-minimal failure set $M \subseteq C$}
$n \leftarrow 2$\;
\While{$|C| \geq 2$}{
  Partition $C$ into $n$ chunks $\Delta_1, \ldots, \Delta_n$\;
  $\mathit{found} \leftarrow \textbf{false}$\;
  \For{$i = 1$ \KwTo $n$}{
    \If{$\mathrm{test}(C \setminus \Delta_i) = \mathrm{FAIL}$}{
      $C \leftarrow C \setminus \Delta_i$\;
      $n \leftarrow \max(n - 1, 2)$\;
      $\mathit{found} \leftarrow \textbf{true}$\;
      \textbf{break}\;
    }
  }
  \If{\textbf{not} $\mathit{found}$}{
    \lIf{$n \geq |C|$}{\textbf{break}}
    $n \leftarrow \min(2n, |C|)$\;
  }
}
\Return{$C$}\;
\end{algorithm}

\subsection{Proxy Oracle for WorkArena}
\label{appendix:proxy_oracle}

In Phase~2 of MFS construction (\S\ref{subsec:mfs_construction}), ddmin iteratively removes subsets of the candidate set $C_s$ and invokes an oracle to test whether the remaining elements still induce task failure.
On WebLinx, this oracle is inexpensive: since the benchmark evaluates single-step action prediction, the oracle simply feeds the modified HTML to the agent and checks whether the step-wise reward of the predicted action falls below $0.5$.
On WorkArena, however, each oracle call would require running the agent from step $s$ to task completion over multiple sequential steps with live web access, which is costly.

To avoid this cost, we define a proxy oracle based on the erroneous action observed in Phase~1.
In the failed rollouts of Phase~1, the agent produced an erroneous action at step $s$ when all self-reported elements $C_s$ were removed.
Two independent full rollouts confirmed that this action leads to task failure.
We thus use this action as the failure criterion in ddmin.
For each test, we feed the modified HTML at step $s$ to the agent.
If the agent produces the same erroneous action, the test counts as FAIL.
This reduces each oracle call to a single inference step.

\subsection{DOM Distance-Based Partitioning for ddmin}

Standard ddmin partitions elements into contiguous chunks by input order, which is arbitrary for DOM elements.
MFS elements are likely to be structurally close in the DOM tree, as task-critical information often concentrates in a localized region of the page.
If such related elements fall into the same chunk, ddmin can eliminate the complementary chunk early and reduce oracle calls.
We therefore define a distance on the DOM tree and partition elements so that nearby elements are grouped together.

To partition elements by DOM proximity, we first define a distance between elements.
Let $e_i = (b_i, a_i)$ and $e_j = (b_j, a_j)$, where $b$ denotes the element identifier and $a$ the attribute name. The distance is:
\begin{equation}
d(e_i, e_j) = \begin{cases}
0 & b_i = b_j,\; a_i = a_j \\
1 & b_i = b_j,\; a_i \neq a_j \\
\text{hop}(b_i, b_j) + 1 & \text{otherwise}
\end{cases}
\end{equation}
where $\text{hop}(b_i, b_j)$ is the number of edges on the path between $b_i$ and $b_j$ through their lowest common ancestor in the DOM tree.
Using this distance, we partition elements via Farthest Point Sampling (FPS): we iteratively select $k$ seed elements, each farthest from the already selected seeds, then greedily assign remaining elements to the nearest seed's group (with a size constraint of $\lceil |C| / k \rceil$ per group).
To evaluate this partitioning, we compare FPS against random partitioning via simulation.
The oracle used in Phase~2 of MFS construction (\S\ref{subsec:mfs_construction}) is non-deterministic, so FPS and random partitioning may find different 1-minimal sets, preventing a fair direct comparison of oracle call counts.
We therefore fix a ground-truth MFS and define a deterministic oracle: $\text{test}(S) = \text{FAIL}$ if $\text{MFS} \subseteq S$, and PASS otherwise.
We evaluate under two settings: Setting~A uses the MFS found by FPS as ground truth (favorable to FPS), and Setting~B uses the MFS found by random partitioning (favorable to random).
Each setting is run 50 times per case and we report the average oracle call count.

Table~\ref{tab:ablation_fps_vs_random} shows the results.
FPS reduces oracle calls by 1.39 on WorkArena and 0.28 on WebLinx under Setting~A, and by 0.19 and 0.15 under Setting~B.
The improvement is modest because the majority of MFS instances consist of a single element (80\% on WorkArena, 71\% on WebLinx), for which any partitioning strategy converges quickly.
FPS consistently outperforms random partitioning under both settings, so we adopt it as the default.

\begin{table}[t]
\centering
\small
\caption{Ablation study: FPS-based vs.\ random partitioning in ddmin.
Average oracle calls over 50 simulation trials per case.
Setting~A uses the MFS found by FPS as ground truth (favorable to FPS);
Setting~B uses the MFS found by random as ground truth (favorable to random).}
\label{tab:ablation_fps_vs_random}
\begin{tabular}{l|rr|rr}
\toprule
 & \multicolumn{2}{c|}{Setting A} & \multicolumn{2}{c}{Setting B} \\
Dataset & FPS & Random & FPS & Random \\
\midrule
WorkArena (n=51) & 7.43 & 8.82 & 7.87 & 8.06 \\
WebLinx (n=41) & 6.05 & 6.33 & 6.32 & 6.47 \\
\bottomrule
\end{tabular}
\end{table}

\section{Additional End-to-End Results}
\label{appendix:additional_e2e}

\subsection{Paired Comparisons of Success Rates}
\label{appendix:paired}

\begin{table*}[t]
\centering
\small
\caption{Paired comparisons of end-to-end success rate between GEPA (ratio=0.2) and each method with Qwen3.5-122B-A10B. $\Delta$ success denotes the success rate of GEPA (ratio=0.2) minus that of the competitor, computed per task from the average over two runs. For each competitor, the setting is the $k$ whose reduction ratio is closest to that of GEPA (ratio=0.2). The $p$-value is from a two-sided paired sign-flip permutation test with 10,000 permutations. The 95\% CI is obtained by bootstrap with 10,000 resamples, where the 33 task types on WorkArena and the 300 instances on WebLinx are resampled with replacement. $^*$ denotes $p < 0.05$.}
\label{tab:paired}
\begin{tabular}{llccc}
\toprule
Competitor & Setting & Reduction ratio & $\Delta$ success [95\% CI] & $p$ \\
\midrule
\multicolumn{5}{l}{\textit{WorkArena L1}} \\
\quad DMR (BM25) & $k=50$ & 0.236 & +0.288 [+0.136, +0.455] & .002$^*$ \\
\quad DMR (Dense) & $k=50$ & 0.259 & +0.121 [$-$0.030, +0.273] & .209 \\
\quad DMR (QueryGen) & $k=50$ & 0.203 & +0.121 [+0.015, +0.242] & .079 \\
\quad Prune4Web & $k=10$ & 0.104 & +0.167 [+0.045, +0.303] & .031$^*$ \\
\quad FocusAgent & $k=50$ & 0.215 & +0.091 [+0.000, +0.197] & .183 \\
\quad Pruned by AXTree & --- & 0.501 & +0.045 [$-$0.076, +0.182] & .660 \\
\quad Original (HTML) & --- & 1.000 & $-$0.076 [$-$0.227, +0.061] & .408 \\
\midrule
\multicolumn{5}{l}{\textit{WebLinx}} \\
\quad DMR (BM25) & $k=10$ & 0.261 & +0.046 [+0.019, +0.076] & .001$^*$ \\
\quad DMR (Dense) & $k=10$ & 0.269 & +0.057 [+0.028, +0.089] & .000$^*$ \\
\quad DMR (QueryGen) & $k=10$ & 0.265 & +0.024 [+0.005, +0.043] & .015$^*$ \\
\quad Prune4Web & $k=10$ & 0.289 & +0.005 [$-$0.018, +0.027] & .656 \\
\quad FocusAgent & $k=10$ & 0.293 & $-$0.002 [$-$0.018, +0.012] & .750 \\
\quad Pruned by AXTree & --- & 0.559 & $-$0.005 [$-$0.021, +0.012] & .598 \\
\quad Original (HTML) & --- & 1.000 & $-$0.017 [$-$0.032, $-$0.002] & .029$^*$ \\
\bottomrule
\end{tabular}

\end{table*}

\begin{table*}[t]
\centering
\small
\caption{End-to-end success rate with bootstrap 95\% confidence intervals for the Original (HTML) and GEPA (ratio=0.2) with Qwen3.5-122B-A10B.}
\label{tab:success_ci}
\begin{tabular}{llccc}
\toprule
Benchmark & Method & Reduction ratio & Success rate [95\% CI] & Latency per step \\
\midrule
WorkArena L1 & Original (HTML) & 1.000 & 0.485 [0.318, 0.652] & 65.7s \\
WorkArena L1 & GEPA (ratio=0.2) & 0.183 & 0.409 [0.242, 0.576] & 30.2s \\
WebLinx & Original (HTML) & 1.000 & 0.144 [0.111, 0.180] & 30.0s \\
WebLinx & GEPA (ratio=0.2) & 0.226 & 0.127 [0.095, 0.161] & 9.7s \\
\bottomrule
\end{tabular}

\end{table*}

\S\ref{sec:comp_comparison/end2end} shows that GEPA (ratio=0.2) matches or exceeds the success rate of every other method at a comparable reduction ratio, while retaining most of the success rate of the Original (HTML).
Because these success rates are measured on a limited number of tasks, we test whether these differences are statistically significant.
We compare GEPA (ratio=0.2) with each method by a paired test over tasks, which cancels the difference in difficulty between tasks.
For each method, we use the $k$ whose reduction ratio is closest to that of GEPA (ratio=0.2).
Table~\ref{tab:paired} shows that GEPA (ratio=0.2) is significantly better in 5 of the 12 comparisons with reduction methods and never significantly worse.
Compared with the Original (HTML), the drop in success rate is not significant on WorkArena.
On WebLinx, the drop is significant: the success rate decreases from 0.144 by 0.017, with a 95\% confidence interval of 0.002 to 0.032 for the decrease.
Thus, GEPA (ratio=0.2) is never worse than the other reduction methods, while on WebLinx its speedup comes with a statistically significant drop from the Original (HTML).
For reference, Table~\ref{tab:success_ci} reports the success rates of GEPA (ratio=0.2) and the Original (HTML) with bootstrap 95\% confidence intervals.

\subsection{Per-Step Latency Decomposition}
\label{appendix:latency_decomp}

\begin{table*}[t]
\centering
\small
\caption{Per-step latency in seconds decomposed into four stages with Qwen3.5-122B-A10B. Preproc denotes observation preprocessing other than HTML reduction, such as HTML parsing. Reduction denotes HTML reduction by the reduction method, including LLM inference for LLM inference-based methods. Policy denotes inference of the agent's policy model. Web env denotes browser operation and environment access, which is negligible on WebLinx because it is evaluated without live web access. Values are averaged over two runs and, for methods with a selection parameter, over $k \in \{10, 50, 100, 200\}$.}
\label{tab:latency_decomp}
\setlength{\tabcolsep}{4pt}
\begin{tabular}{l|ccccc|ccccc}
\toprule
 & \multicolumn{5}{c|}{WorkArena L1} & \multicolumn{5}{c}{WebLinx} \\
Method & Preproc & Reduction & Policy & Web env & Total & Preproc & Reduction & Policy & Web env & Total \\
\midrule
Original (HTML) & 0.26 & 0.00 & 47.30 & 18.17 & 65.7 & 0.12 & 0.00 & 29.75 & 0.08 & 30.0 \\
GEPA (ratio=0.2) & 0.25 & 0.11 & 11.68 & 18.18 & 30.2 & 0.12 & 0.10 & 9.39 & 0.08 & 9.7 \\
Pruned by AXTree & 0.25 & 0.19 & 16.87 & 17.53 & 34.8 & 0.13 & 0.03 & 20.29 & 0.08 & 20.5 \\
DMR (BM25) & 0.27 & 0.11 & 13.44 & 16.78 & 30.6 & 0.13 & 0.03 & 14.31 & 0.09 & 14.6 \\
DMR (Dense) & 0.27 & 2.85 & 13.80 & 17.27 & 34.2 & 0.13 & 0.87 & 14.07 & 0.08 & 15.2 \\
DMR (QueryGen) & 0.26 & 12.75 & 9.20 & 17.16 & 39.4 & 0.13 & 12.76 & 7.25 & 0.08 & 20.2 \\
Prune4Web & 0.25 & 54.13 & 7.84 & 17.46 & 79.7 & 0.13 & 46.91 & 5.64 & 0.08 & 52.8 \\
FocusAgent & 0.24 & 186.25 & 6.41 & 17.60 & 210.5 & 0.13 & 184.55 & 4.92 & 0.08 & 189.7 \\
\bottomrule
\end{tabular}

\end{table*}

To locate where the per-step latency in \S\ref{sec:comp_comparison/end2end} is spent, we decompose it into four stages: observation preprocessing other than HTML reduction, such as HTML parsing; HTML reduction; policy model inference; and web environment access (Table~\ref{tab:latency_decomp}).
The speedup of GEPA (ratio=0.2) comes from the shorter input to the policy model, while its own reduction takes about 0.1s.
On WorkArena, its policy model inference takes 11.68s per step, compared with 47.30s for the Original (HTML).
For LLM inference-based methods, the reduction step itself takes most of the per-step latency, which offsets their faster policy model inference.
For example, FocusAgent spends 186.25s of its 210.5s per step on reduction on WorkArena.

\subsection{Results with MiniMax-M2.5}
\label{appendix:minimax}

\begin{figure*}[t]
  \centering
  \includegraphics[width=\linewidth]{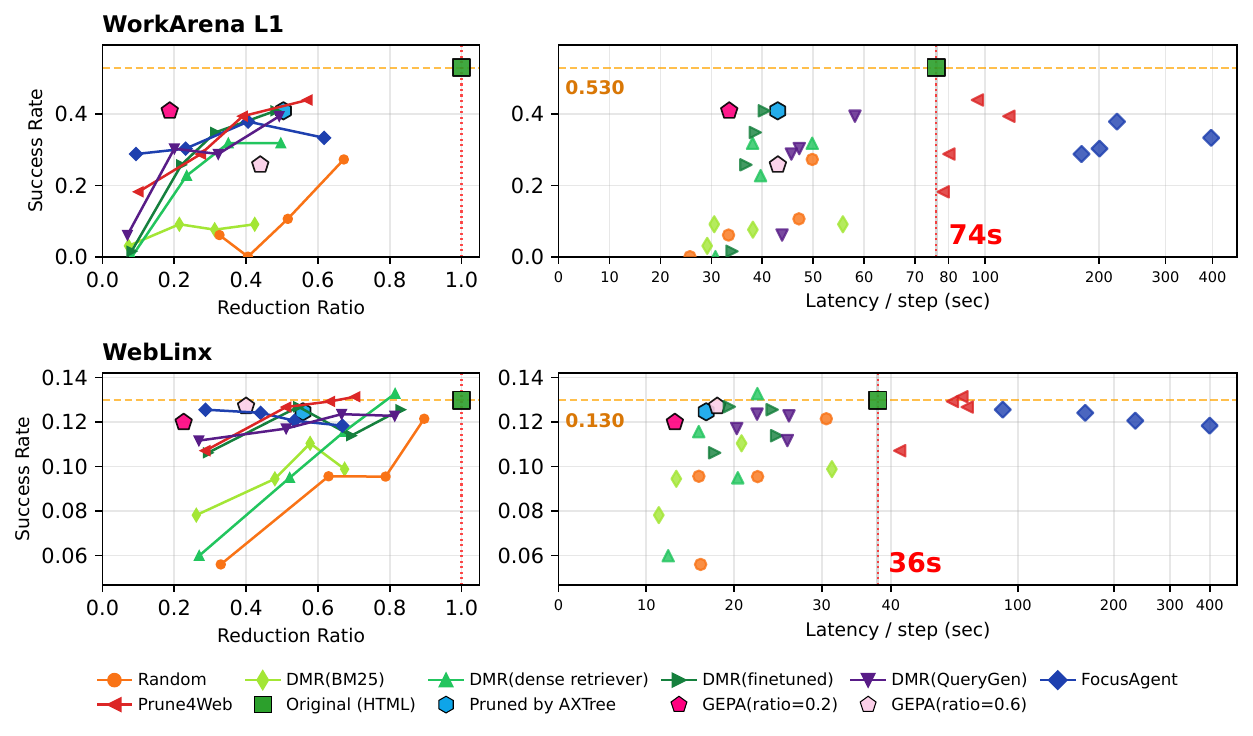}
  \caption{End-to-end evaluation with MiniMax-M2.5 as the policy model. Left column: success rate versus reduction ratio. Right column: success rate versus average wall-clock latency per step. The vertical red dashed line indicates the latency of passing the full HTML without reduction. The horizontal orange dashed line indicates the success rate of passing the full HTML without reduction. The first row shows results on WorkArena L1 and the second row on WebLinx.}
  \label{fig:e2e_minimax}
\end{figure*}

Figure~\ref{fig:e2e_minimax} shows the end-to-end evaluation results using MiniMax-M2.5 as the policy model. The overall trends are consistent with the Qwen3.5-122B-A10B results presented in \S\ref{sec:comp_comparison/end2end}.

\section{Case Studies}
\label{appendix:case_studies}

We present two MFS instances from the coverage evaluation on WorkArena L1: one in which GEPA retains the MFS while retrieval-based methods miss it, and one in which GEPA itself fails.
In the HTML below, \texttt{bid} is the element identifier.

\noindent\textbf{Case 1: GEPA retains the MFS and retrieval-based methods miss it.}
In the \texttt{create-user} task, the agent must set the Date format field at step 5 by \texttt{select\_option('a405', 'dd/MM/yyyy')} on the following HTML.
\begin{lstlisting}
<select bid="a405" id="sys_user.date_format" class="form-control" ...>
  <option bid="a408" value="dd/MM/yyyy">dd/MM/yyyy</option>
  ...
</select>
\end{lstlisting}
The MFS consists of four entries: the tag and the \texttt{id} attribute of the \texttt{select} element, and the \texttt{value} attribute and the text of the \texttt{option} element.
GEPA (ratio=0.2) retains all four, because its learned program keeps interactive tags such as \texttt{select} and \texttt{option} and keeps the \texttt{id} and \texttt{value} attributes (Appendix~\ref{appendix:gepa_programs}).
DMR (BM25) misses them even at $k=500$, and DMR (Dense) retains them only at $k \geq 100$, because the text \texttt{dd/MM/yyyy} shares no word with the task goal.

\noindent\textbf{Case 2: GEPA fails.}
In the \texttt{filter-\allowbreak incident-\allowbreak list} task, the agent configures a list filter.
At step 1, the correct action is \texttt{select\_option('a1141', 'assignment\_group')}.
The MFS is a single entry, the \texttt{class} attribute of another \texttt{select} element, \texttt{a1176}.
\begin{lstlisting}
<select bid="a1176" class="filerTableSelect select2 form-control filter_type ..." ...>
\end{lstlisting}
This element's class names appear to be the cue in the HTML that \texttt{a1141} is the element to operate on, since removing this attribute alone makes the agent fail.
GEPA (ratio=0.2) keeps the element itself because \texttt{select} is an interactive tag.
However, it removes \texttt{class}, which is not on its attribute keep-list.
Without this cue, the agent clicks a wrong control, \texttt{click('a1166')}, which is a ``-- choose field --'' link, instead of issuing the correct \texttt{select\_option} action.
DMR (Dense) and FocusAgent succeed on this instance because they keep whole elements and do not prune attributes.
The optimization did not acquire a rule covering this case from its training instances.

\section{GEPA Pruning Programs}
\label{appendix:gepa_programs}

Listing~\ref{lst:gepa_seed} shows the seed program that defines the initial pruning strategy from which GEPA's evolution begins.
Listings~\ref{lst:gepa_wa02} and~\ref{lst:gepa_wl02} show the learned programs for WorkArena ($r{=}0.2$) and WebLinx ($r{=}0.2$), respectively.

\begin{lstlisting}[caption={GEPA seed pruning program.},label={lst:gepa_seed},style=python]
from bs4 import BeautifulSoup, Tag
import re, functools

try:
    from nltk.stem import PorterStemmer
    _STEMMER = PorterStemmer()
    STEM_AVAILABLE = True
except ImportError:
    STEM_AVAILABLE = False

def prune_html(html, task_goal, action_history):
    soup = BeautifulSoup(html, "html.parser")
    for tag_name in ["head", "script", "style",
                     "link", "meta"]:
        for tag in soup.find_all(tag_name):
            tag.extract()

    @functools.lru_cache(maxsize=None)
    def stem(word):
        if STEM_AVAILABLE:
            return _STEMMER.stem(word.lower())
        return word.lower()

    query = f"{task_goal} {action_history}".lower()
    stemmed_kw = set(stem(w)
        for w in re.findall(r"\w+", query)
        if len(w) > 2)

    interactive = {"input", "button", "select",
                   "textarea", "a", "label", "option"}
    keep = set()
    for el in soup.find_all(attrs={"bid": True}):
        if not isinstance(el, Tag):
            continue
        bid = el["bid"]
        if el.name in interactive:
            keep.add(bid); continue
        text = el.get_text(separator=" ",
                           strip=True).lower()
        tokens = {stem(w)
            for w in re.findall(r"\w+", text)}
        if tokens & stemmed_kw:
            keep.add(bid)

    for el in soup.find_all(attrs={"bid": True}):
        if isinstance(el, Tag) and el["bid"] in keep:
            for p in el.parents:
                if isinstance(p, Tag) \
                   and p.has_attr("bid"):
                    keep.add(p["bid"])

    for el in list(soup.find_all(attrs={"bid": True})):
        if isinstance(el, Tag) \
           and el["bid"] not in keep:
            el.extract()
    return str(soup)
\end{lstlisting}

\begin{lstlisting}[caption={GEPA learned program: WorkArena ($r{=}0.2$).},label={lst:gepa_wa02},style=python]
from bs4 import BeautifulSoup, Tag, NavigableString
import re, functools
from collections import deque

try:
    from nltk.stem import PorterStemmer
    _STEMMER = PorterStemmer()
    STEM_AVAILABLE = True
except ImportError:
    STEM_AVAILABLE = False

ATTRIBUTES_TO_KEEP = {
    "bid", "id", "name", "value", "type",
    "href", "src", "alt", "title", "placeholder",
    "aria-label", "data-label", "for", "role",
    "checked", "selected", "disabled", "readonly"}

def prune_html(html, task_goal, action_history):
    soup = BeautifulSoup(html, "html.parser")
    for t in ["head","script","style","link","meta"]:
        for tag in soup.find_all(t): tag.extract()

    @functools.lru_cache(maxsize=None)
    def stem(word):
        if STEM_AVAILABLE:
            return _STEMMER.stem(word.lower())
        return word.lower()

    query = f"{task_goal} {action_history}".lower()
    stemmed_kw = set(stem(w)
        for w in re.findall(r"\w+", query)
        if len(w) > 1)

    action_bids = set()
    select_targets = set()
    for m in re.finditer(
        r"(?:click|fill)\('([^']+)'\)"
        r"|select_option\('([^']+)',\s*'([^']+)'\)",
        action_history):
        if m.group(1): action_bids.add(m.group(1))
        elif m.group(2) and m.group(3):
            action_bids.add(m.group(2))
            select_targets.add(m.group(3))

    bid_map = {el["bid"]: el
        for el in soup.find_all(attrs={"bid": True})}
    interactive = {"input","button","select",
        "textarea","a","label","option"}
    keep = set()

    for el in bid_map.values():
        bid = el["bid"]
        if bid in action_bids: keep.add(bid)
        if el.name == "option" and select_targets:
            v = el.get("value","")
            t = el.get_text(strip=True)
            if v in select_targets \
               or t in select_targets:
                keep.add(bid)
        texts = [el.get_text(separator=" ",strip=True)]
        for a in ["title","alt","aria-label",
                   "placeholder","value","data-label"]:
            if el.has_attr(a): texts.append(el[a])
        combined = " ".join(filter(None,texts)).lower()
        tokens = {stem(w)
            for w in re.findall(r"\w+", combined)
            if len(w) > 1}
        if el.name in interactive \
           or (tokens & stemmed_kw):
            keep.add(bid)

    q = deque(list(keep)); seen = set(keep)
    while q:
        el = bid_map.get(q.popleft())
        if el:
            for p in el.parents:
                if isinstance(p, Tag):
                    if p.name == "body": break
                    if p.has_attr("bid") \
                       and p["bid"] not in seen:
                        keep.add(p["bid"])
                        seen.add(p["bid"])
                        q.append(p["bid"])

    for el in list(bid_map.values()):
        if isinstance(el, Tag) \
           and el["bid"] not in keep:
            el.extract()

    for par in list(soup.find_all(attrs={"bid":True})):
        if isinstance(par, Tag) \
           and par["bid"] in keep:
            for ch in list(par.contents)[::-1]:
                if isinstance(ch, Tag):
                    if not ch.has_attr("bid"):
                        ct = ch.get_text(separator=" ",
                            strip=True)
                        ts = {stem(w)
                            for w in re.findall(
                                r"\w+", ct.lower())
                            if len(w) > 1}
                        if not (ts & stemmed_kw):
                            ch.extract()
                elif isinstance(ch, NavigableString):
                    c = re.sub(r'\s+',' ',
                        str(ch)).strip()
                    if not c: ch.extract()
                    else: ch.replace_with(c)
                else: ch.extract()

    for el in list(soup.find_all(attrs={"bid":True})):
        if isinstance(el, Tag):
            el.attrs = {k: v
                for k, v in el.attrs.items()
                if k.lower() in ATTRIBUTES_TO_KEEP}

    for tn in list(soup.find_all(string=True)):
        if isinstance(tn, NavigableString):
            c = re.sub(r'\s+',' ',str(tn)).strip()
            if not c: tn.extract()
            else: tn.replace_with(c)
    return str(soup)
\end{lstlisting}

\begin{lstlisting}[caption={GEPA learned program: WebLinx ($r{=}0.2$).},label={lst:gepa_wl02},style=python]
from bs4 import BeautifulSoup, Tag, NavigableString
import re, functools

try:
    from nltk.stem import PorterStemmer
    _STEMMER = PorterStemmer()
    STEM_AVAILABLE = True
except ImportError:
    STEM_AVAILABLE = False

GLOBAL_PRESERVED_ATTRIBUTES = {
    "bid", "id", "name", "value", "type",
    "href", "src", "alt", "title", "placeholder",
    "aria-label", "role", "checked", "selected",
    "disabled", "contenteditable", "for",
    "colspan", "rowspan", "tabindex", "maxlength",
    "data-testid", "data-test-id", "data-test",
    "content"}

TEXTUAL_RELEVANCE_ATTRIBUTES = {
    "value", "placeholder", "aria-label",
    "title", "alt", "name", "content"}

def prune_html(html, task_goal, action_history):
    soup = BeautifulSoup(html, "html.parser")
    for t in ["script", "style", "noscript"]:
        for tag in soup.find_all(t): tag.extract()

    @functools.lru_cache(maxsize=None)
    def stem(word):
        if STEM_AVAILABLE:
            return _STEMMER.stem(word.lower())
        return word.lower()

    query = f"{task_goal} {action_history}".lower()
    stemmed_kw = set(stem(w)
        for w in re.findall(r"\w+", query)
        if len(w) > 2)

    all_bid = list(
        soup.find_all(attrs={"bid": True}))
    keep = set()
    interactive = {"input","button","select",
        "textarea","a","label","option"}

    for el in all_bid:
        if not isinstance(el, Tag): continue
        bid = el["bid"]
        if el.name in interactive:
            keep.add(bid); continue
        if el.has_attr("contenteditable") \
           and el["contenteditable"] == "true":
            keep.add(bid); continue
        if el.name == "title":
            keep.add(bid); continue
        if el.name == "meta" \
           and el.get("name") == "description":
            keep.add(bid); continue

        parts = []
        for ch in el.contents:
            if isinstance(ch, NavigableString):
                s = str(ch).strip()
                if s: parts.append(s)
        for a in TEXTUAL_RELEVANCE_ATTRIBUTES:
            if el.has_attr(a):
                v = el[a]
                if isinstance(v, list):
                    parts.extend(v)
                else: parts.append(str(v))
        text = " ".join(
            p for p in parts if p).lower()
        if text:
            tokens = {stem(w)
                for w in re.findall(r"\w+", text)}
            if tokens & stemmed_kw:
                keep.add(bid)

    keep_final = set(keep)
    for el in all_bid:
        if el["bid"] in keep_final \
           and el.has_attr("contenteditable") \
           and el["contenteditable"] == "true":
            for d in el.find_all(attrs={"bid":True}):
                keep_final.add(d["bid"])

    bid_map = {el["bid"]: el for el in all_bid}
    kept_tags = set(bid_map[b]
        for b in keep_final if b in bid_map)

    new_soup = BeautifulSoup(
        features="html.parser")

    def _build(node, parent):
        if isinstance(node, NavigableString):
            s = str(node).strip()
            if not s: return
            orig = node.parent
            if isinstance(orig, Tag) and (
                orig.get("bid") in keep_final
                or orig.name in ["html","body"]):
                if parent is not None:
                    parent.append(
                        NavigableString(node))
            return
        if not isinstance(node, Tag): return

        is_kept = node in kept_tags
        is_root = node.name in ["html", "body"]
        target = parent

        if is_kept or is_root:
            t = new_soup.new_tag(node.name)
            for a, v in node.attrs.items():
                if a in GLOBAL_PRESERVED_ATTRIBUTES:
                    t[a] = v
            if parent is not None:
                parent.append(t)
            else:
                new_soup.append(t)
            target = t

        for ch in node.contents:
            _build(ch, target)

    for ch in soup.contents:
        _build(ch, new_soup)

    if not new_soup.contents:
        return "<html><body></body></html>"
    return str(new_soup)
\end{lstlisting}

\section{DOM Normalization for Trajectory Replay}
\label{appendix:dom_normalization}

In the intervention experiments on WorkArena\footnote{WebLinx operates on single-step evaluation and does not require trajectory replay or DOM normalization.} (\S\ref{subsec:mfs_construction}), each test requires reproducing the state at step $s$ by replaying recorded actions on a fresh browser session before applying the intervention.
Each recorded action refers to its target by an element identifier, but the DOM is not identical across sessions because of dynamic content such as timestamps and elements that depend on load timing.
We therefore normalize the DOM of both sessions, renumber the element identifiers sequentially, and match elements with the same identifier.
The normalization function was developed iteratively: Claude Code\footnote{\url{https://code.claude.com/docs/en/overview}} was used to analyze DOM differences across sessions and propose candidate heuristics, and the authors validated each rule to ensure it only removes session-dependent variation.
The complete set of normalization rules is listed below, organized by category.

\paragraph{Category A: Session-specific identifiers and dynamic UI elements.}\mbox{}\par
\noindent\textbf{A1.} Remove system announcement banner elements (content varies by timing).

\noindent\textbf{A2.} Remove application scope and update set info panels.

\noindent\textbf{A3.} Normalize page titles in \texttt{<title>} to \texttt{[PAGE\_TITLE]}.

\noindent\textbf{A4.} Normalize experience title spans to \texttt{[PAGE\_TITLE]}.

\noindent\textbf{A5.} Normalize favorite button \texttt{aria-label} containing page title to \texttt{[PAGE\_TITLE]}.

\noindent\textbf{A6.} Remove dynamically initialized search and typeahead components (initialization-dependent structure).

\noindent\textbf{A7.} Replace URL cache-buster parameters with placeholder.

\noindent\textbf{A8.} Normalize auto-numbered record identifiers (e.g., incident, change, problem, and request prefixes) with placeholders.

\noindent\textbf{A9.} Sort dynamically ordered variable map items by a stable key (non-deterministic ordering).

\noindent\textbf{A10.} Normalize Highcharts dynamic IDs and timing-dependent values.

\noindent\textbf{A11.} Remove browser-automation attributes from empty screen-reader-only spans (variable presence by timing).

\noindent\textbf{A12.} Delete empty screen-reader-only spans entirely.

\noindent\textbf{A13.} Remove per-session component identifier attributes.

\noindent\textbf{A14.} Remove dynamic CSS classes related to animation, visibility, and placeholder state.

\noindent\textbf{A15.} Remove resulting empty \texttt{class=""} and \texttt{style=""} attributes.

\noindent\textbf{A16.} Remove miscellaneous dynamic attributes (directionality, truncation handlers, tooltip references, animation markers).

\noindent\textbf{A17.} Remove dynamically added \texttt{style} from \texttt{<iframe>} elements.

\noindent\textbf{A18.} Normalize position-related layout attributes.

\noindent\textbf{A19.} Remove SVG inner content (asynchronously loaded icon paths).

\noindent\textbf{A20.} Remove page initialization script lines.

\noindent\textbf{A21.} Normalize dynamic score display title attributes.

\noindent\textbf{A22.} Remove the page timing widget (response time display).

\noindent\textbf{A23.} Remove audio elements (notification sounds).

\noindent\textbf{A24.} Remove empty \texttt{<style>} tags.

\noindent\textbf{A25.} Remove context menu configuration script blocks.

\noindent\textbf{A26.} Clear content of screen-reader-only spans (dynamic notification messages).

\noindent\textbf{A27.} Remove content of \texttt{aria-live} containers (dynamic status messages).

\noindent\textbf{A28.} Remove Select2 hidden-accessible spans.

\noindent\textbf{A29.} Normalize Select2 auto-generated IDs.

\noindent\textbf{A30.} Remove dynamically generated Bootstrap tooltips.

\noindent\textbf{A31.} Remove asynchronously loaded home page widget elements.

\noindent\textbf{A32.} Remove list table rows (database-state dependent).

\noindent\textbf{A33.} Remove hidden main-content fallback structure.

\noindent\textbf{A34.} Normalize tooltip component dynamic attributes (\texttt{aria-hidden}, \texttt{aria-label}, \texttt{role}).

\noindent\textbf{A35.} Remove tooltip host container elements.

\noindent\textbf{A36.} Remove BrowserGym marker attributes.

\noindent\textbf{A37.} Remove randomly generated framework-internal attributes.

\noindent\textbf{A38.} Normalize relative time displays (e.g., \texttt{5m ago} and \texttt{N from now}~$\to$~\texttt{[TIMEAGO]}).

\noindent\textbf{A39.} Remove empty \texttt{value=""} attributes set dynamically by JavaScript.

\paragraph{Category B: \texttt{<font>} tag conversion and CSS normalization.}\mbox{}\par
\noindent\textbf{B1.} Convert deprecated \texttt{<font>} tags to \texttt{<span style="...">}, mapping \texttt{size} to \texttt{font-size} and \texttt{face} to \texttt{font-family}.

\noindent\textbf{B2.} Sort CSS properties alphabetically within \texttt{<span>} \texttt{style} attributes to absorb ordering differences.

\paragraph{Category C: Dynamic IDs and layout values.}\mbox{}\par
\noindent\textbf{C1.} Replace UUIDs with \texttt{[UUID]}.

\noindent\textbf{C2.} Remove loading spinner elements.

\noindent\textbf{C3.} Remove natural-language-query help modal elements.

\noindent\textbf{C4.} Remove function field modal elements.

\noindent\textbf{C5.} Remove widget category listing elements.

\noindent\textbf{C6.} Replace 32-character hex platform identifiers with \texttt{[SYS\_ID]}.

\noindent\textbf{C7.} Remove load-completion indicator elements.

\noindent\textbf{C8.} Remove floating chat container elements.

\noindent\textbf{C9.} Normalize floating collaboration container dynamic attributes.

\noindent\textbf{C10.} Normalize ARIA reference IDs (\texttt{aria-controls}, \texttt{aria-labelledby}, heading IDs, etc.) by replacing dynamic portions with placeholders.

\noindent\textbf{C11.} Normalize sequential ID parts in dropdown lists and triggers.

\noindent\textbf{C12.} Normalize TinyMCE auto-generated ARIA IDs.

\noindent\textbf{C13.} Normalize tooltip numeric IDs.

\noindent\textbf{C14.} Normalize dynamically numbered window identifiers and compiled content.

\noindent\textbf{C15.} Normalize dynamically generated section IDs and GridStack layout IDs.

\noindent\textbf{C16.} Renumber element identifiers (bid) sequentially within each frame, and keep the mapping between the original and renumbered identifiers.

\paragraph{Category D: User and instance identity.}\mbox{}\par
\noindent\textbf{D1.} Extract user display name from multiple DOM sources (JavaScript variables, ARIA labels, text content) and replace all occurrences with \texttt{[USER\_NAME]}, \texttt{[USER\_FIRST]}, \texttt{[USER\_LAST]}, including initials and account patterns.

\noindent\textbf{D2.} Normalize instance-specific URLs.

\noindent\textbf{D3.} Replace email addresses matching platform patterns with \texttt{[USER\_EMAIL]}.

\noindent\textbf{D4.} Normalize category rendering component IDs.

\noindent\textbf{D5.} Normalize dashboard container IDs.

\noindent\textbf{D6.} Normalize content slot identifier values.

\noindent\textbf{D7.} Normalize welcome messages.

\noindent\textbf{D8.} Replace icon image paths with \texttt{[ICON\_IMAGE]}.

\noindent\textbf{D9.} Normalize SVG \texttt{viewbox} values.

\paragraph{Category E: Session tokens and server state.}\mbox{}\par
\noindent\textbf{E1.} Replace CSRF token values with \texttt{[SESSION\_TOKEN]}.

\noindent\textbf{E2.} Normalize user identity JavaScript variables.

\noindent\textbf{E3.} Normalize server-side time values in JavaScript variables.

\noindent\textbf{E4.} Normalize URL timestamps.

\noindent\textbf{E5.} Normalize dynamic \texttt{height} in style attributes.

\noindent\textbf{E6.} Normalize theme hash parameters.

\noindent\textbf{E7.} Replace encoded list property values with placeholder.

\noindent\textbf{E8.} Replace navigation history JSON with placeholder.

\noindent\textbf{E9.} Normalize user session object in JavaScript globals.

\noindent\textbf{E10.} Normalize encoded record parameter values in URLs.

\noindent\textbf{E11.} Normalize referring URL parameters (user names, times, and encoded data within back-link URLs).

\noindent\textbf{E12.} Replace client-side cache-buster values.

\noindent\textbf{E13.} Replace client-side session state JSON.

\noindent\textbf{E14.} Normalize page performance metric values.

\noindent\textbf{E15.} Normalize session debug information.

\paragraph{Category F: Database-dependent display values.}\mbox{}\par
\noindent\textbf{F1.} Normalize row count displays (\texttt{Showing rows N to N of ...}).

\noindent\textbf{F2.} Normalize row count data attributes.

\noindent\textbf{F3.} Replace dates (\texttt{YYYY-MM-DD}) and times (\texttt{HH:MM:SS}) with placeholders.

\noindent\textbf{F4.} Normalize record count URL parameters.

\noindent\textbf{F5.} Normalize CSS position values.

\noindent\textbf{F6.} Remove empty notification placeholder spans.

\noindent\textbf{F7.} Remove browsing context hidden input elements.

\noindent\textbf{F8.} Remove catalog option price annotations.

\noindent\textbf{F9.} Normalize record update timestamps in \texttt{data-updated-on} attributes to \texttt{[DATETIME]}.

\paragraph{Category G: Whitespace normalization.}\mbox{}\par
Trailing whitespace and empty lines are removed, and standalone numeric lines are replaced with \texttt{[ROW\_COUNT]}.

\end{document}